\documentclass[conference]{IEEEtran}
\IEEEoverridecommandlockouts
\usepackage{cite}
\usepackage{amsmath,amssymb,amsfonts}
\usepackage{algorithmic}
\usepackage{graphicx}
\usepackage{textcomp}
\usepackage{xcolor}
\usepackage{subfig}
\usepackage{todonotes}
\usepackage{multirow}
\usepackage{floatrow}
\usepackage{tabu}
\usepackage{color, colortbl}
\def\BibTeX{{\rm B\kern-.05em{\sc i\kern-.025em b}\kern-.08em
    T\kern-.1667em\lower.7ex\hbox{E}\kern-.125emX}}
\usepackage{threeparttable}
    
\newcommand{\ineq}[1]{\footnotesize$#1$\normalsize}{}
\newcommand{\tech}{\text{{PyCARL}}}{}    
\newcommand{\carlsim}{\text{{CARLsim}}}{} 
\usepackage{xargs}
\newcommand{\nc}[1]{\textcolor{black}{#1}}
\newcommandx{\jeffNote}[2][1=]{\todo[inline,linecolor=black,backgroundcolor=orange!25,bordercolor=orange,#1]{#2 ---Jeff}}

\newcommandx{\hkNote}[2][1=]{\todo[inline,linecolor=black,backgroundcolor=blue!25,bordercolor=blue,#1]{#2 ---Hirak}}

\newcommandx{\NikNote}[2][1=]{\todo[inline,linecolor=black,backgroundcolor=yellow!25,bordercolor=yellow,#1]{#2 ---Nik}}
\newcommandx{\adNote}[2][1=]{\todo[inline,linecolor=black,#1]{#2 ---Anup}}
    
\begin{document}

% \title{\Small PyCARL: A Common PyNN Interface for GPU-Accelerated Biologically Plausible \\Spiking Neural Network Simulation\\
% % {\footnotesize \textsuperscript{*}Note: Sub-titles are not captured in Xplore and
% % should not be used}
% % \thanks{Identify applicable funding agency here. If none, delete this.}
% }
% \title{\Small \tech{}: An Integrated PyNN Interface for Spiking Neural Network Simulation and Hardware Mapping}
\title{\tech{}: A PyNN Interface for Hardware-Software Co-Simulation of Spiking Neural Network}

\author{\IEEEauthorblockN{Adarsha Balaji${}^1$, Prathyusha Adiraju${}^2$, Hirak J. Kashyap${}^3$, Anup Das${}^{1,2}$, \\Jeffrey L. Krichmar${}^3$, Nikil D. Dutt${}^3$, and Francky Catthoor${}^{2,4}$}
\IEEEauthorblockA{\textit{${}^1$Electrical and Computer Engineering, Drexel University, Philadelphia, USA} \\
\textit{\textit{${}^2$Neuromorphic Computing, Stichting Imec Nederlands, Eindhoven, Netherlands}} \\
\textit{\textit{${}^3$Cognitive Science and Computer Science, University of California, Irvine, USA}} \\
\textit{\textit{${}^4$ESAT Department, KU Leuven and IMEC, Leuven, Belgium}} \\
Correspondence Email: anup.das@drexel.edu, jkrichma@uci.edu, Francky.Catthoor@imec.be},
}

% \and
% \and
% \and
% \and
% \and

\maketitle

\begin{abstract}
We present \tech{}, a PyNN-based common Python programming interface for hardware-software co-simulation of spiking neural network (SNN).
Through \tech{}, we make the following two key contributions.
First, we provide an interface of PyNN to \carlsim{}, a computationally-efficient, GPU-accelerated and biophysically-detailed SNN simulator.
\tech{} facilitates joint development of machine learning models and code sharing between CARLsim and PyNN users, promoting an integrated and larger neuromorphic community.
\nc{Second, we integrate cycle-accurate models of state-of-the-art neuromorphic hardware such as TrueNorth, Loihi, and DynapSE in \tech{},
to accurately model hardware latencies, which delay spikes between communicating neurons, degrading performance of machine learning models.}
\tech{} allows users to analyze and optimize the performance difference between software-based simulation and hardware-oriented simulation. We show that system designers can also use \tech{} to perform design-space exploration early in the product development stage, facilitating faster time-to-market of neuromorphic products.
\end{abstract}


\section{Introduction}\label{sec:introduction}
Advances in computational neuroscience have produced a variety of software for simulating spiking neural network (SNN)~\cite{maass1997networks} ---
%, which facilitates machine learning application development with neurobiological details. 
%Examples of these simulators include 
NEURON \cite{hines1997neuron}, NEST \cite{gewaltig2007nest}, PCSIM \cite{pecevski2009pcsim}, Brian~\cite{goodman2008brian}, MegaSim~\cite{megasim}, and \carlsim{}~\cite{chou2018carlsim}. These simulators model neural functions at various levels of detail and therefore have different requirements for computational resources.

\nc{In this paper, we focus on \carlsim{}~\cite{chou2018carlsim}, 
%the latest version of the SNN simulation library CARLsim~\cite{nageswaran2009configurable}, 
which facilitates} parallel simulation of large SNNs using CPUs and multi-GPUs, simulates multiple compartment models, 9-parameter Izhikevich and leaky integrate-and-fire (LIF) spiking neuron models, and integrates the fourth order Runge Kutta (RK4) method for improved numerical precision. CARLsim's support for built-in biologically realistic neuron, synapse, current and emerging learning models and continuous
integration and testing, make it an easy to use and powerful simulator of biologically-plausible SNN models. Benchmarking results demonstrate simulation of 8.6 million neurons and 0.48 billion synapses using 4 GPUs and
up to 60x speedup with multi-GPU implementations over a single-threaded CPU implementation.%, making \carlsim{}
%well-suited for simulating large-scale SNN models~\cite{chou2018carlsim}.

%\adNote{Do we know the \# of CARLsim users?}

To facilitate faster application development and portability across research institutes, a common Python programming interface called PyNN is proposed~\cite{davison2009pynn}.
PyNN provides a high-level abstraction of SNN models, promotes code sharing and reuse, and provides a foundation for simulator-agnostic analysis, visualization and data-management tools. 
Many SNN simulators now support interfacing with PyNN --- PyNEST~\cite{eppler2009pynest} for the NEST simulator,
PyPCSIM~\cite{pecevski2009pcsim} for the PCSIM simulator,
and Brian~2~\cite{marcel2019brian} for the Brian simulator.
\nc{Through PyNN, applications developed using one simulator can be analyzed/simulated using another simulator with minimal effort.}

%\noindent\textbf{Contribution 1:} 
\nc{Currently, no interface exists between \carlsim{}, which is implemented in C++ and the Python-based PyNN. Therefore, applications developed in PyNN cannot be analyzed using \carlsim{} and conversely, \carlsim{}-based applications cannot be analyzed/simulated using other SNN simulators without requiring significant effort.} This creates a large gap between these two research communities.

\begin{figure}[h!]
	\centering
	\vspace{-5pt}
	\centerline{\includegraphics[width=0.99\columnwidth]{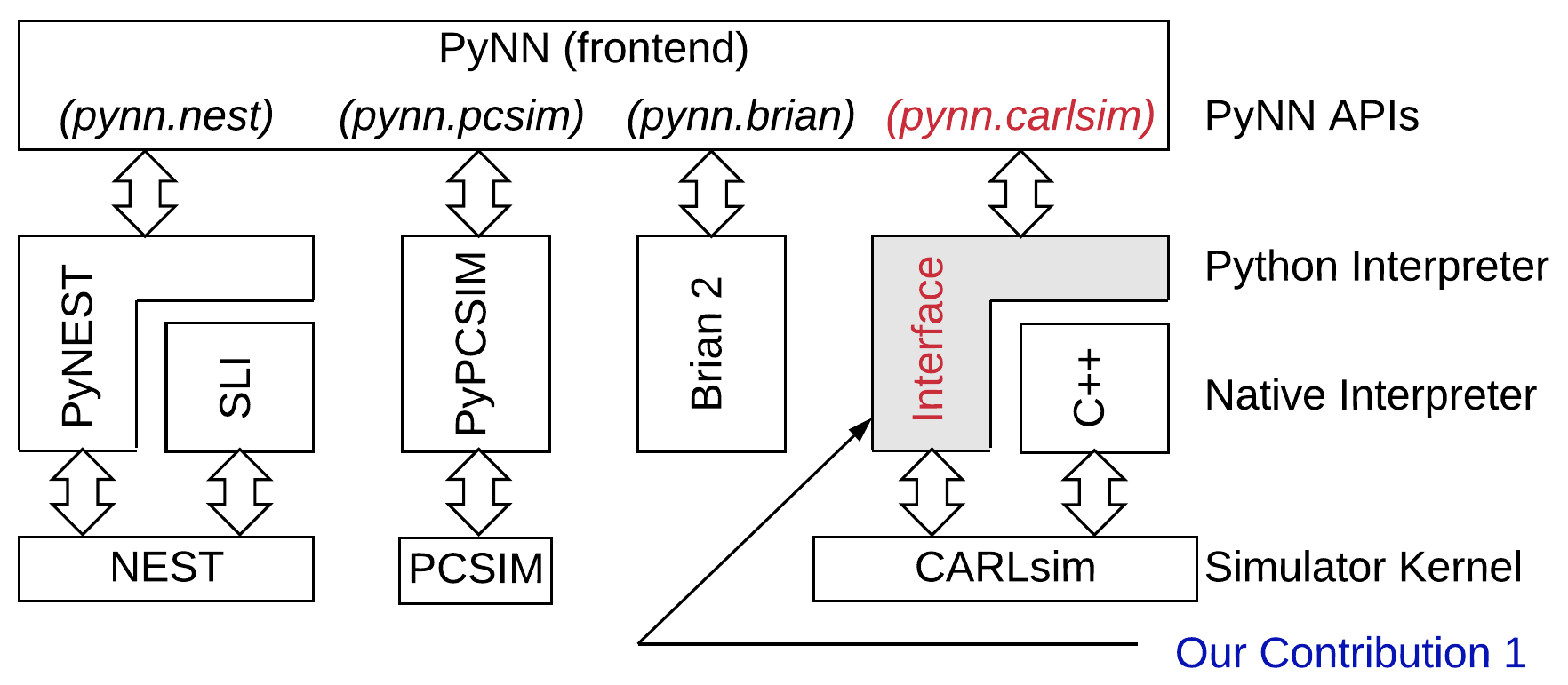}}
	%\vspace{-5pt}
	\caption{PyNN standardized API architecture and our proposed pynn-to-carlsim interface.}%Python interface for \carlsim{}.}
	\vspace{-10pt}
	\label{fig:contribution_1}
\end{figure}

Our \textbf{objective} is to bridge this gap and create an integrated neuromorphic research community, facilitating joint developments of machine learning models and efficient code sharing.
Figure~\ref{fig:contribution_1} illustrates the standardized application programming interface (API) architecture in PyNN. Brian~2 and PCSIM, which are native Python implementations, employ
a direct communication via the \texttt{pynn.brian} and \texttt{pynn.pcsim} API calls, respectively.
NEST, on the other hand, is not a native Python simulator. So, the \texttt{pynn.nest} API call first results in a code generation to the native SLI code, a stack-based language derived from PostScript. The generated code is then used by the Python interpreter PyNEST to simulate an SNN application utilizing the backend NEST simulator kernel.
%interpreter that utilizes the NEST kernel backend for SNN simulation.
%Other simulators such as NEST and PCSIM have a different native
%interpreter; an additional Python interpreter is designed to interface with
%PyNN.
%These simulators can be used either through their respective native
%interpreters or through the Python interpreter exposed in the PyNN API. 
Figure \ref{fig:contribution_1} also shows our proposed interface for \carlsim{}, which is exposed via the
new \texttt{pynn.carlsim} API in PyNN. We describe this interface in details in Section \ref{sec:simulation}.% and analyze its performance in Sec.~\ref{sec:results}.

%\vspace{-10pt}
%\jeffNote{Contribution 2 reads more like a result rather than an objective.  Need to state what the contribution is at the outset. What is the hardware used in Fig. 2. Was it developed by Drexel or IMEC? Does it have a name?}

% \jeffNote{Doesn't the following paragraph belong with Contribution 2?}
On the hardware front, neuromorphic computing~\cite{mead1990neuromorphic} has shown significant promise to fuel the growth of machine learning, thanks to low-power design of neuron circuits, distributed implementation of computing and storage,
%in the form of crossbars\footnote{A crossbar is a two dimensional organization of row and column wires with a synaptic element such as non-volatile memory at each crosspoint.}, 
and integration of non-volatile synaptic memory. In recent years, several spiking neuromorphic architectures are designed: SpiNNaker~\cite{furber2014spinnaker}, DYNAP-SE~\cite{Moradi2018ADYNAPs}, TrueNorth~\cite{debole2019truenorth} and Loihi \cite{davies2018loihi}. 
\nc{Unfortunately, due to non-zero latency of hardware components, spikes between communicating neurons may experience non-deterministic delays, impacting SNN performance.}

\vspace{-5pt}

\begin{figure}[h!]
	\centering
	\vspace{-5pt}
	\centerline{\includegraphics[width=0.9\columnwidth]{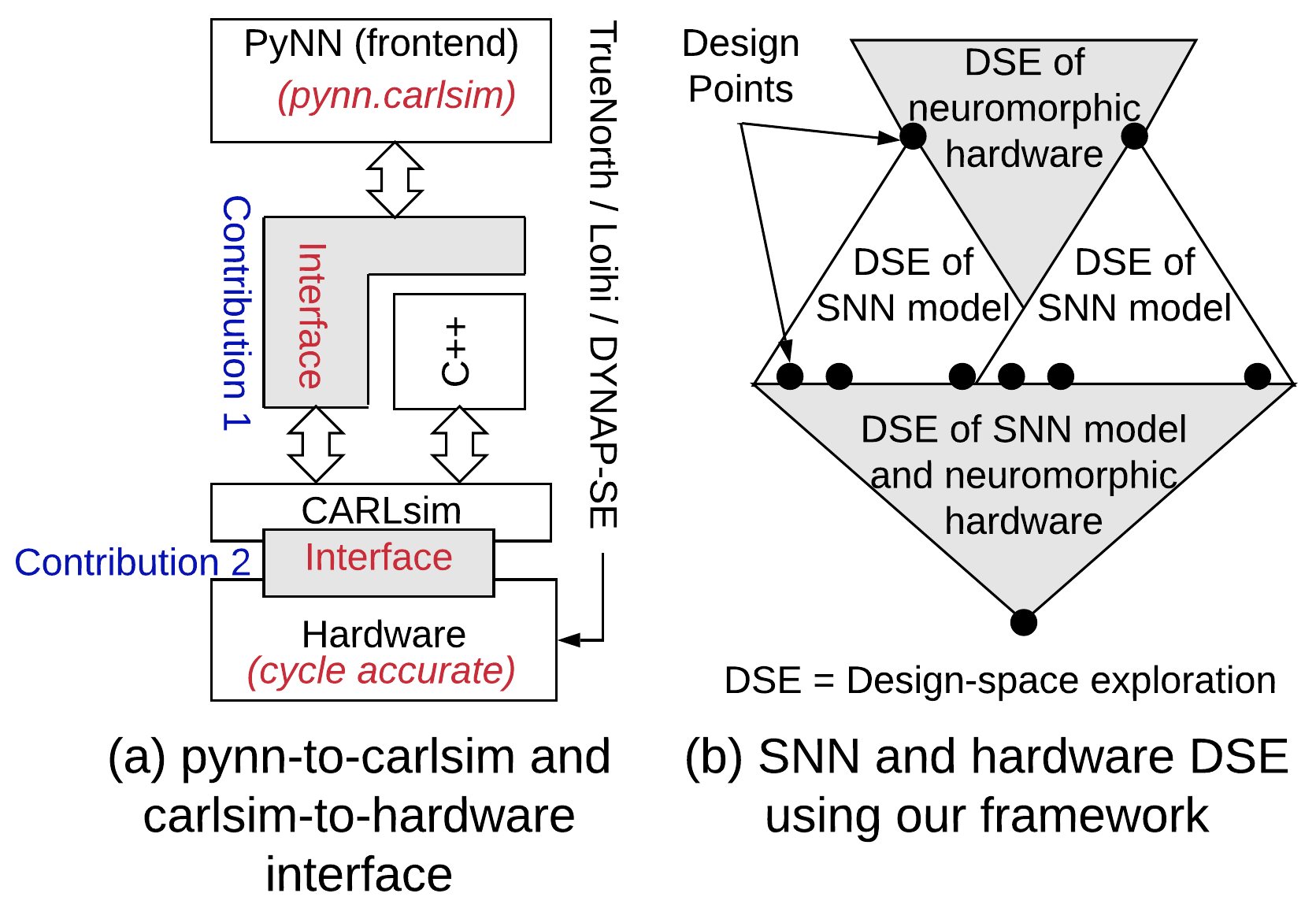}}
	%\vspace{-10pt}
	\caption{(a) Our proposed interface to estimate SNN performance on neuromorphic hardware and (b) design space exploration (DSE) based on this contribution.}
	\vspace{-10pt}
	\label{fig:contribution_2a}
\end{figure}

%\noindent\textbf{Contribution 2:} 
\nc{Currently, no PyNN-based simulators incorporate neuromorphic hardware laterncies.}
Therefore, SNN performance estimated using PyNN can be different from the performance obtained on hardware. Our \textbf{objective} is to estimate this performance difference, allowing users to optimize their machine learning model \nc{to meet a desired performance on a target neuromorphic hardware.} Figure \ref{fig:contribution_2a} shows our proposed \texttt{carlsim-to-hardware} interface to model state-of-the-art neuromorphic hardware at a cycle-accurate level, using the output generated from the proposed \texttt{pynn-to-carlsim} interface (see Figure \ref{fig:contribution_1}). We describe this interface in Sec.~\ref{sec:hardware_simulation}.

\vspace{-5pt}

\nc{The two new interfaces developed in this work can be integrated inside a design-space exploration (DSE) framework (illustrated in Figure \ref{fig:contribution_2a}(b)) to explore different SNN topologies and neuromorphic hardware configurations, optimizing both SNN performance such as accuracy and hardware performance such as latency, energy, and throughput.}

\noindent\textbf{Summary:} To summarize, our comprehensive co-simulation framework, which we call \tech{}, allows \carlsim{}-based detailed software simulations, hardware-oriented simulations, and neuromorphic design-space explorations, all from a common PyNN frontend, allowing extensive portability across different research institutes. By using cycle-accurate models of state-of-the-art neuromorphic hardware, \tech{} allows users to perform hardware exploration and performance estimation
%, without the necessity to procure these hardware platforms 
early during application development, accelerating the neuromorphic product development cycle.

\section{Our Integrated Framework \tech{}}\label{sec:framework}
\begin{figure}[h!]
	\centering
	\vspace{-5pt}
	\centerline{\includegraphics[width=0.9\columnwidth]{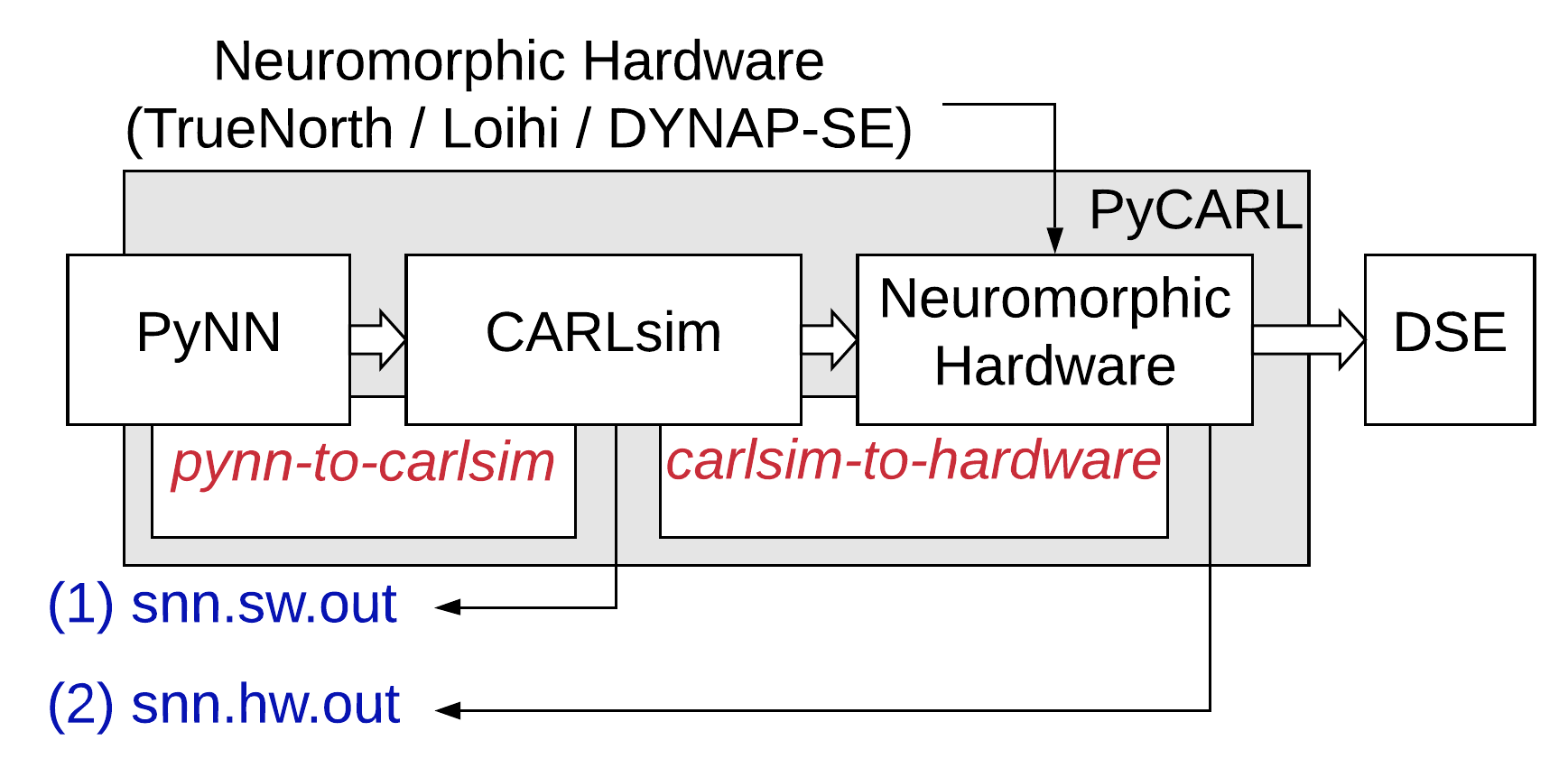}}
	\caption{Our integrated framework \tech{}.}
	\vspace{-10pt}
	\label{fig:integrated_simulator}
\end{figure}

Figure \ref{fig:integrated_simulator} shows a high-level overview of our integrated framework \tech{}, based on PyNN. An SNN model written in PyNN is simulated using the \carlsim{} backend kernel with the proposed  pynn-to-carlsim interface (contribution 1). This generates the first output \textcolor{blue}{\texttt{snn.sw.out}}, which consists of synaptic strength of each connection in the network and precise timing of spikes on these connections. This output is then used in the proposed carlsim-to-hardware interface, allowing simulating the SNN on a cycle-accurate model of a state-of-the-art neuromorphic hardware such as TrueNorth \cite{debole2019truenorth}, Loihi \cite{davies2018loihi}, and DYNAP-SE \cite{Moradi2018ADYNAPs}. Our cycle-accurate model generates the second output \textcolor{blue}{\texttt{snn.hw.out}}, which consists of 1) hardware-specific metrics such as latency, throughput, and energy, and 2) SNN-specific metrics such as inter-spike interval distortion and disorder spike count (which we formulate and elaborate in Section \ref{sec:hardware_simulation}). SNN-specific metrics estimate the performance drop due to non-zero hardware latencies.

%This spike information, called \texttt{spike trace}, is next used for hardware-software co-simulation using a cycle-accurate simulator, which models neuromorphic hardware such as TrueNorth~\cite{debole2019truenorth}, Loihi~\cite{davies2018loihi}, and DynapSE~\cite{Moradi2018ADYNAPs}. 
%This step generates the second output -- \textcolor{blue}{\emph{snn.hw.out}}, which consists of spike information incorporating latencies of the given neuromorphic hardware.
%The hardware-aware spike information is next used for design-space exploration to optimize for energy, latency, or any other metric on the neuromorphic hardware. This step generates the final output -- \textcolor{blue}{\emph{snn.map.out}}, which consists of information regarding the mapping of neurons and synapses of the SNN to crossbars of the neuromorphic hardware.

\vspace{-5pt}

\nc{We now describe the components of \tech{} and show how to use \tech{} to perform design space explorations.}
% and the data structures used to implement them.

\section{Pynn-to-carlsim Interface in \tech{}}\label{sec:simulation}
Apart from bridging the gap between the PyNN and the CARLsim research communities, the proposed pynn-to-carlsim interface is also significant in the following three ways. First, Python being an interactive language allows users to interact with the \carlsim{} kernel through command line, reducing the application development time. Second, the proposed interface allows code portability across different operating systems (OSes) such as Linux, Solaris, Windows, and Macintosh. Third, Python being open source, allows distributing the proposed interface with mainstream OS releases, exposing neuromorphic computing to the systems community.
%\subsection{Motivation}
%Figure \ref{fig:integrated_simulator} shows a high-level overview of our integrated framework \tech{}, based on PyNN. An SNN application designed in PyNN is simulated in \carlsim{} by using our proposed \texttt{pynn.carlsim} interface. This generates the first output -- \textcolor{blue}{\emph{snn.sw.out}}, which consists of precise timing information of spikes on every synapse of the SNN. This spike information, called \texttt{spike trace}, is next used for hardware-software co-simulation using a cycle-accurate simulator, which models neuromorphic hardware such as TrueNorth~\cite{debole2019truenorth}, Loihi~\cite{davies2018loihi}, and DynapSE~\cite{Moradi2018ADYNAPs}. 
% This step generates the second output -- \textcolor{blue}{\emph{snn.hw.out}}, which consists of spike information incorporating latencies of the given neuromorphic hardware.
% The hardware-aware spike information is next used for design-space exploration to optimize for energy, latency, or any other metric on the neuromorphic hardware. This step generates the final output -- \textcolor{blue}{\emph{snn.map.out}}, which consists of information regarding the mapping of neurons and synapses of the SNN to crossbars of the neuromorphic hardware.

%Of the two approaches to creating an interface between Python (PyNN) and C++ (\carlsim{}), we propose to 

An interface between Python (PyNN) and C++ (\carlsim{}) can be created using the following two approaches. First, through statically linking the C++ library with a Python interpreter. This involves copying all library modules used in \carlsim{} into a final executable image by an external program called linker or link editors. Statically linked files are significantly larger in size because external programs are built into the executable files, which must be loaded into the memory every time they are invoked. This increases program execution time. Static linking also requires all files to be recompiled every time one or more of the shared modules change.
A second approach is the dynamic linking, which involves placing the names of the external libraries (shared libraries) in the final executable file while the actual linking taking place at run time. Dynamic linking is performed by the OS through API calls. Dynamic linking places only one copy of the shared library in memory. This significantly reduces the size of executable programs, thereby saving memory and disk space. Individual shared modules can be updated and recompiled, without compiling the entire source code again. Finally, load time of shared libraries is reduced if the shared library code is already present in memory. Due to lower execution time, reduced memory usage, and flexibility, we adopt dynamic linking of \carlsim{} with PyNN.

%This approach involves compiling the \carlsim{} source code at the user end with every new release of the \carlsim{} simulator. Second, through creation of a dynamic link library and invoking it using an API. This approach provides flexibility by 
%not requiting the user to compile the code, every time a new 
%version of \carlsim{} is released. Of these two approaches, we therefore adopt the dynamic approach of interfacing \carlsim{} with PyNN.

We now describe the two steps involved in creating the proposed pynn-to-carlsim interface.
\subsection{{Step 1}: Generating the Interface Binary \texttt{carlsim.so}}
Unlike PyNEST, which generates the interface binary manually, we propose to use the Simplified Wrapper Interface Generator (SWIG), downloadable at \textcolor{purple}{{http://www.swig.org}}. SWIG simplifies the process of interfacing high level languages such as Python with low-level languages such as C/C++, preserving the robustness and expressiveness of these low-level languages from the high-level abstraction.

\begin{figure}[h!]
	\centering
	\vspace{-5pt}
	\centerline{\includegraphics[width=0.99\columnwidth]{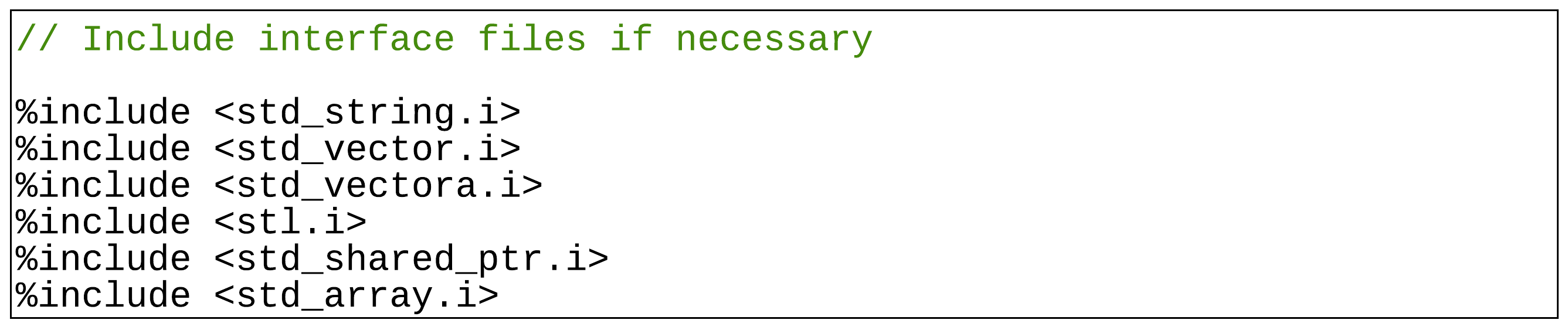}}
	\caption{Define interface files using the \texttt{\%include} directive.}
	\vspace{-10pt}
	\label{fig:swig_interface_file_1}
\end{figure}

\begin{figure}[h!]
	\centering
	\vspace{-10pt}
	\centerline{\includegraphics[width=0.99\columnwidth]{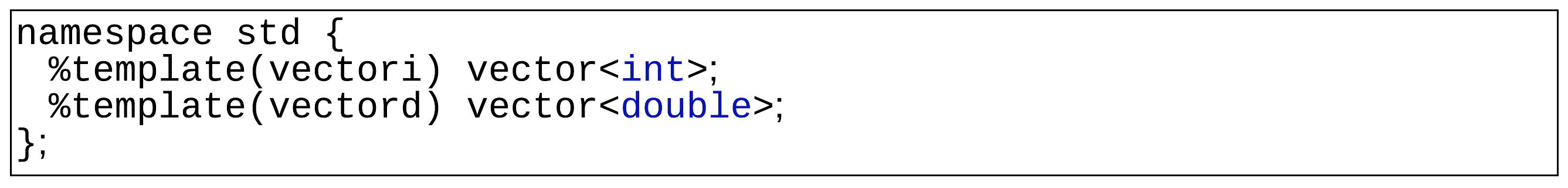}}
	\caption{Declare \carlsim{} data structures using the \texttt{\%template} directive.}
	\vspace{-10pt}
	\label{fig:swig_interface_file_2}
\end{figure}

\begin{figure}[h!]
	\centering
	\vspace{-10pt}
	\centerline{\includegraphics[width=0.99\columnwidth]{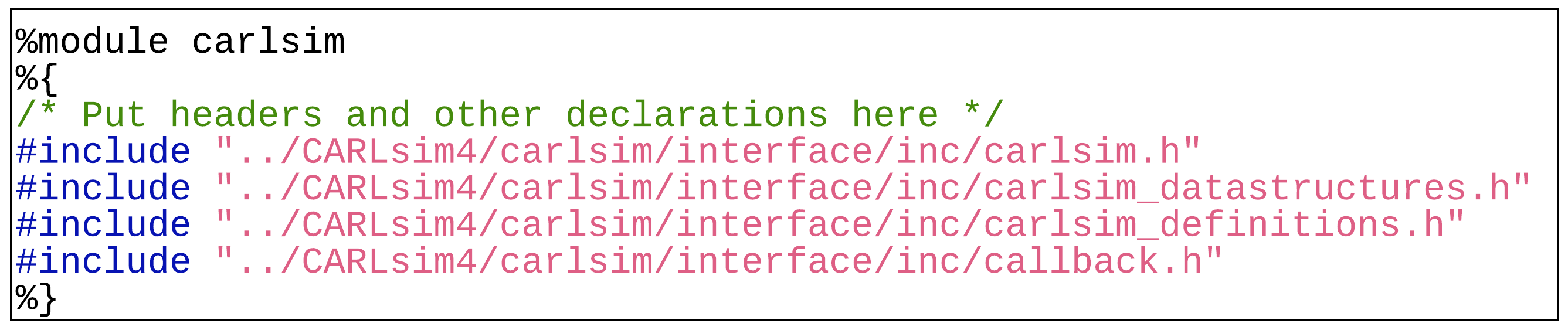}}
	\caption{Main module definition for import in Python.}
	\vspace{-10pt}
	\label{fig:swig_interface_file_3}
\end{figure}

\begin{figure}[h!]
	\centering
	\vspace{-5pt}
	\centerline{\includegraphics[width=0.99\columnwidth]{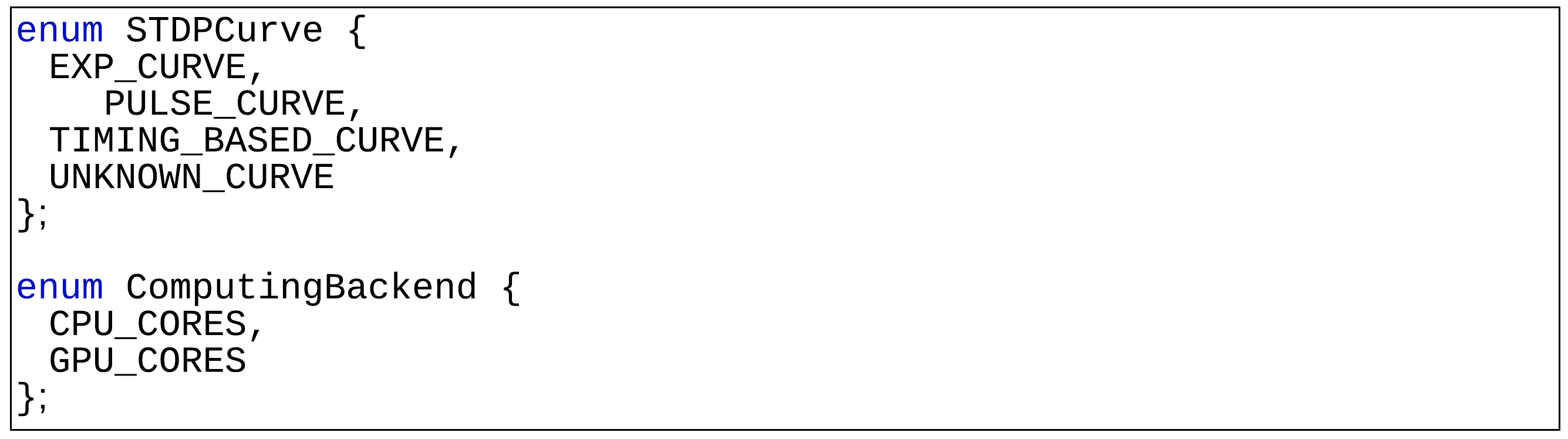}}
	\caption{Enumerated data types using the \texttt{enum} directive.}
	\vspace{-10pt}
	\label{fig:swig_interface_file_4}
\end{figure}

\begin{figure}[h!]
	\centering
	\vspace{-5pt}
	\centerline{\includegraphics[width=0.99\columnwidth]{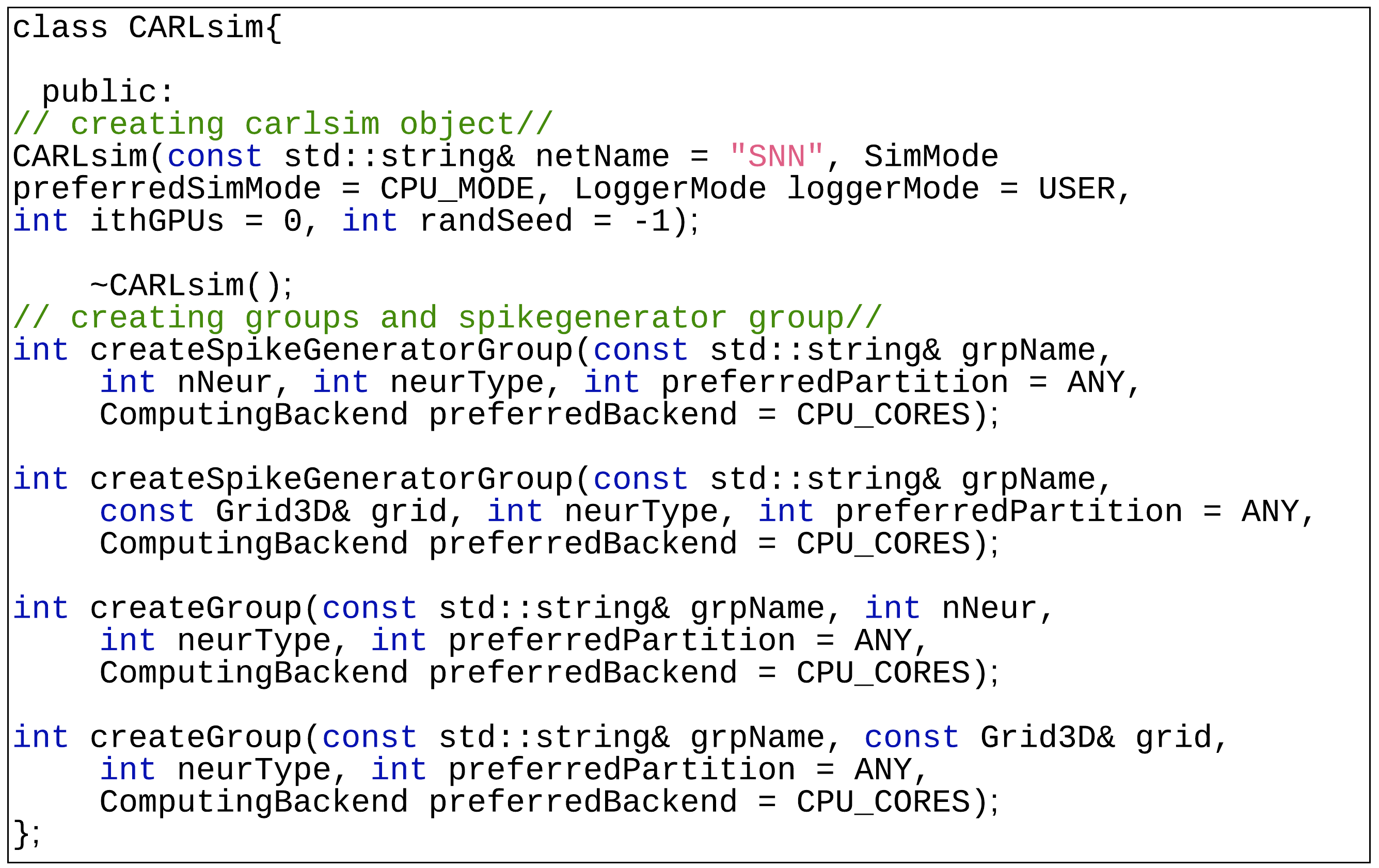}}
	\caption{\carlsim{} class object.}
	\vspace{-10pt}
	\label{fig:swig_interface_file_5}
\end{figure}

The SWIG compiler creates a wrapper binary code by using headers, directives, macros, and declarations from the underlying C++ code of \carlsim{}. Figures \ref{fig:swig_interface_file_1}-\ref{fig:swig_interface_file_5} show the different components of the input file \texttt{carlsim.i} needed to generate the compiled interface binary file \texttt{carlsim.so}. The first component are the interface files that are included using the \texttt{\%include} directive (Figure \ref{fig:swig_interface_file_1}). The second component consists of declaration of the data structures (e.g., vectors) of \carlsim{} using the \texttt{\%template} directive (Figure \ref{fig:swig_interface_file_2}).
The third component is the main module definition that can be loaded in Python using the \texttt{import} command
%. This is shown in 
(Figure \ref{fig:swig_interface_file_3}).
%with the \texttt{\%module} directive containing pointers to the base class and function definitions.

%\jeffNote{Code snippets are hard to read. Make the font larger.}

\vspace{-10pt}

\vspace{-10pt}

The fourth component consists of enumerated data types defined by the directive \texttt{enum} (Figure \ref{fig:swig_interface_file_4}). In this example we show two definitions -- 1) the STDP curve and 2) the computing platform.
The last component is the \carlsim{} class object along with its member functions (Figure \ref{fig:swig_interface_file_5}).

%The SWIG compiler generates a set mapping of {structs} and {constants} in a class definition into constants with the class name as a prefix.

\vspace{-10pt}

\vspace{-10pt}

\begin{figure}[h!]
	\centering
	\vspace{-5pt}
	\centerline{\includegraphics[width=0.99\columnwidth]{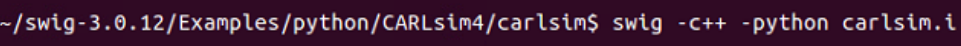}}
	\caption{Compilation of \texttt{carlsim.i} using the SWIG compiler to generate \texttt{carsim.so} interface binary.}
	\vspace{-10pt}
	\label{fig:swig_compilation_step}
\end{figure}

The major advantage of using SWIG is that it uses a layered approach to generate a wrapper over C++ classes. At the lowest level, a collection of procedural ANSI-C style wrappers are generated by SWIG. These wrappers take care of the basic type conversions, type checking, error handling and other low-level details of C++ bindings.
To generate the interface binary file \texttt{carlsim.so}, the input file \texttt{carsim.i} is compiled using the swig compiler as shown in Figure \ref{fig:swig_compilation_step}.

\vspace{-10pt}

\subsection{{Step 2}: Designing PyNN API to Link \texttt{carlsim.so}}
We now describe the proposed \texttt{pynn.carlsim} API to link the interface binary \texttt{carlsim.so} in PyNN.

The \texttt{carlsim.so} interface binary is placed within the sub-package directory of PyNN. This exposes \carlsim{} internal methods as a Python library using the \texttt{import} command as \texttt{\emph{\textcolor{blue}{from carlsim import *}}}. The PyNN front-end API architecture supports implementing both basic functionalities (common for all backend simulators) and specialized simulator-specific functionalities. 
\subsubsection{Implementing common functionalities}
\begin{figure}[h!]
	\centering
	\vspace{-5pt}
	\centerline{\includegraphics[width=1.00\columnwidth]{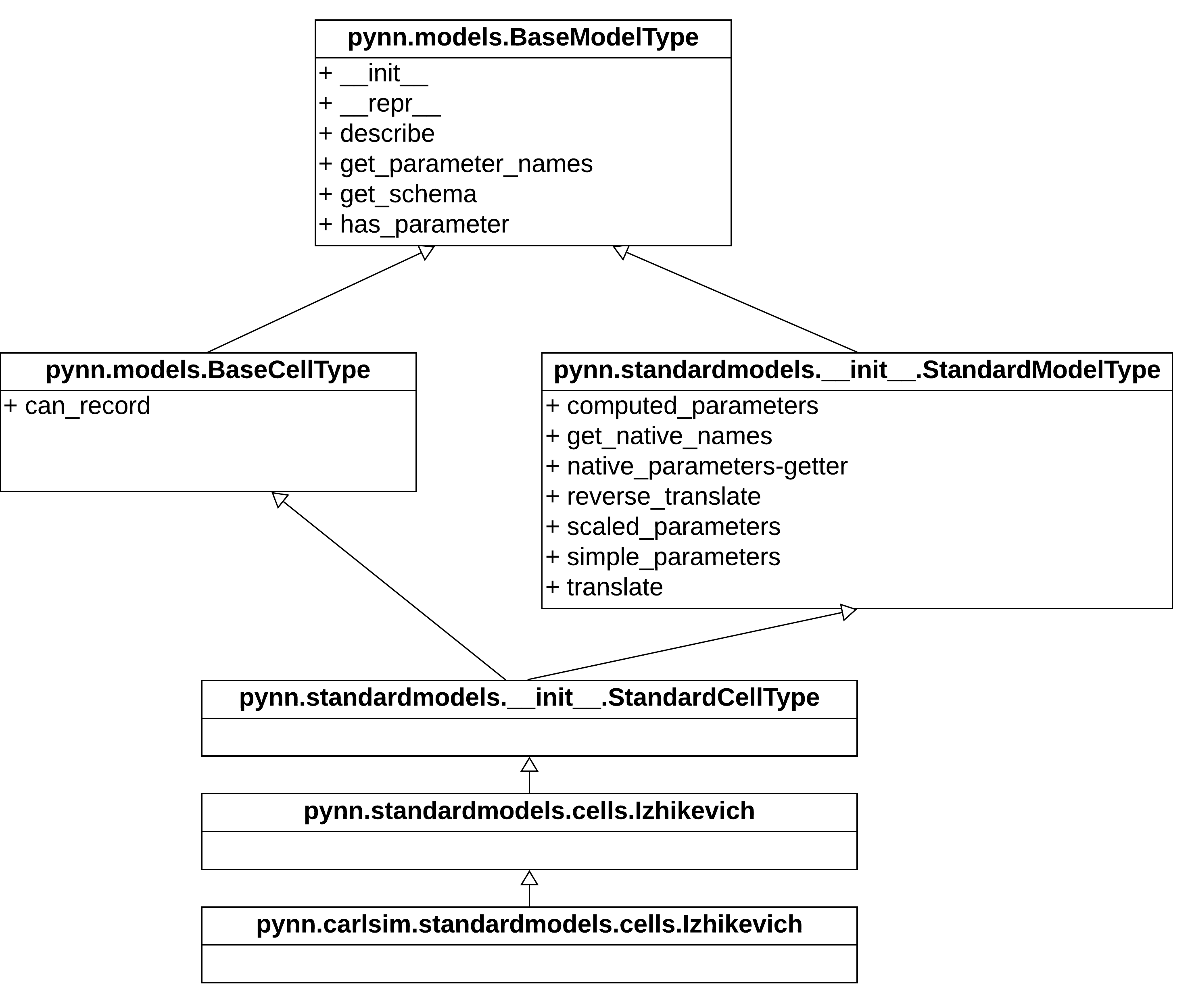}}
	\caption{UML class diagram of Izhikevich cell type showing the relationship with the \texttt{pynn.carlsim} API.}
	\vspace{-10pt}
	\label{fig:create_neuron_model}
\end{figure}

PyNN defines many common functionalities to create a basic SNN model. Examples include cell types, connectors, synapses, and electrodes. Figure \ref{fig:create_neuron_model} shows the UML class diagram to create the Izhikevich cell type \cite{izhikevich2003simple} using the \texttt{pynn.carlsim} API. The inheritance relationship shows that the PyNN \texttt{standardmodels} class includes the definition of all the methods under the \texttt{StandardModelType} class. The Izhikevich model and all similar standard cell types are a specialization of
this \texttt{StandardModelType} class, which subsequently inherits from the PyNN \texttt{BaseModelType} class. Defining other standard components of an SNN model  follow similar inheritance pattern using the common internal API functions provided by PyNN.

\vspace{-10pt}

\subsubsection{Implementing specialized \carlsim{} functions} 
Using the standard PyNN classes, it is also possible to define and expose non-standard \carlsim{} functionalities.

\begin{figure}[h!]
	\centering
	\vspace{-5pt}
	\centerline{\includegraphics[width=0.99\columnwidth]{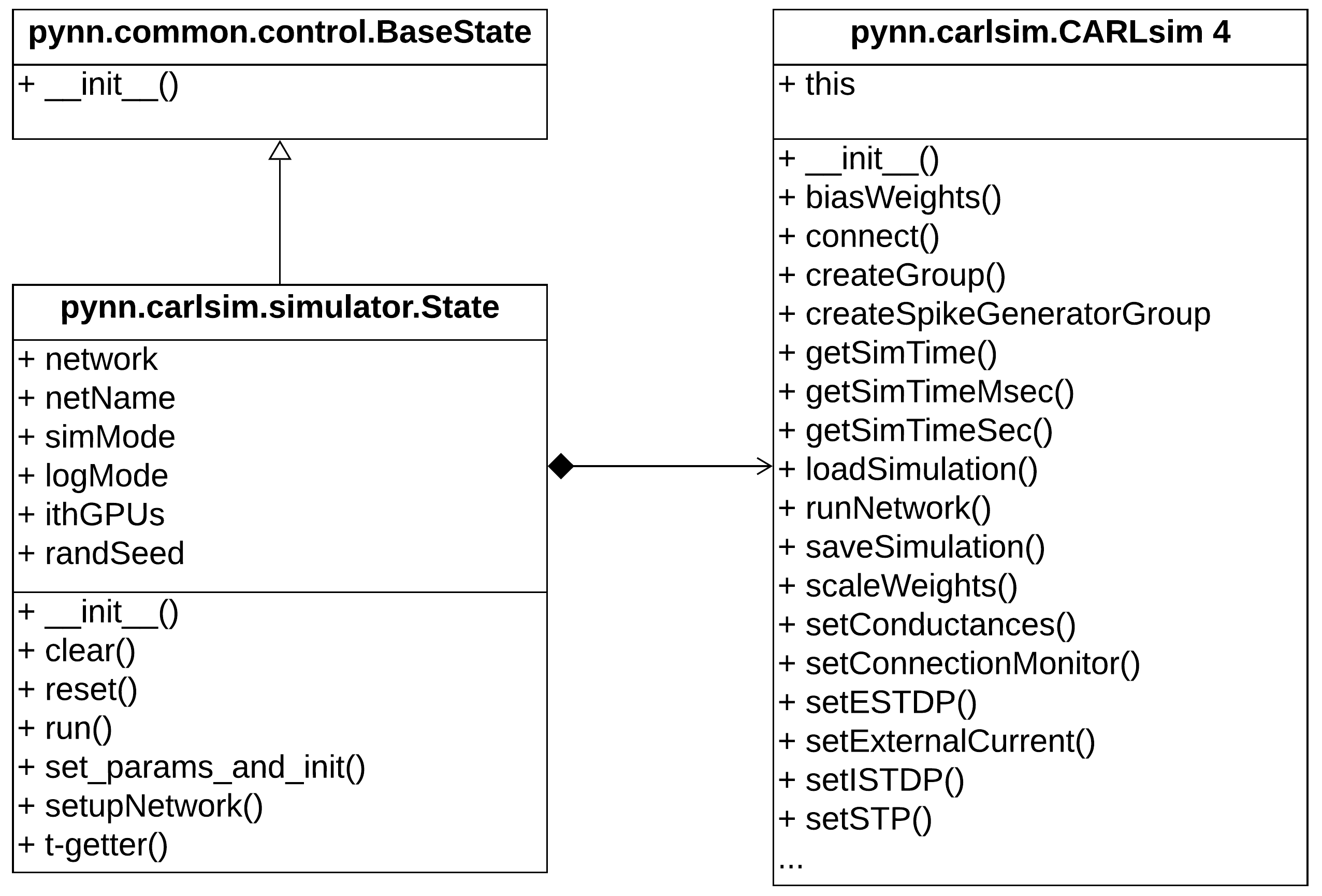}}
	\caption{UML class diagram of simulator state of the \texttt{pynn.carlsim} API.}
	\vspace{-10pt}
	\label{fig:new_functions}
\end{figure}

Figure \ref{fig:new_functions} details the \texttt{state} class of \carlsim{}. The composition adornment
relationship between the \texttt{state} class of the simulator module and the \texttt{\carlsim{}} class in the \texttt{pynn.carlsim} API. The composition
adornment means that apart from the composition relationship between
the contained class (\texttt{\carlsim{}}) and the container class (\texttt{State}), the object of the
contained class also goes out of scope when the containing class goes out of scope.
Thus, the \texttt{State} class exercises complete control over the members of the \texttt{\carlsim{}} class objects. The class member variable \texttt{network} of the \texttt{simulator.State} class contains an instance of the \texttt{\carlsim{}} object.
From Figure \ref{fig:new_functions} it can be seen that
the \texttt{\carlsim{}} class consists of methods which can be used for the initial
configuration of an SNN model and also methods for running, stopping and
saving the simulation. These functions are appropriately exposed to the PyNN 
by including them in the \texttt{pynn.carlsim} API methods, which are created as members
of the class \texttt{simulator.State}.

\begin{figure}[h!]
	\centering
	\vspace{-5pt}
	\centerline{\includegraphics[width=0.99\columnwidth]{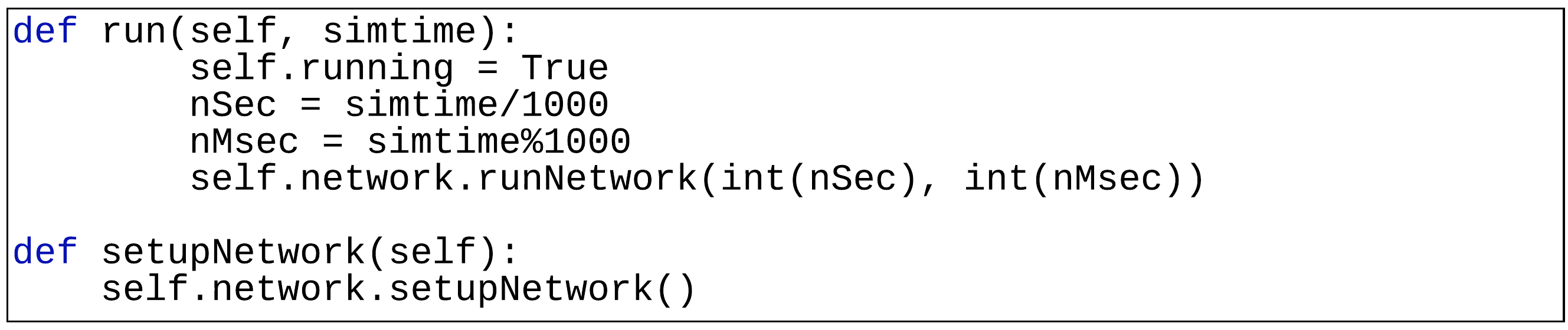}}
	\caption{Snippet showing the exposed \carlsim{} functions in the \texttt{pynn.carlsim} API.}
	\vspace{-10pt}
	\label{fig:define_functions}
\end{figure}

Figure \ref{fig:define_functions} shows the implementations of the \texttt{run()} and \texttt{setupNetwork()} methods in the \texttt{simulator.State} class. It can be seen that these methods call the corresponding functions in the \texttt{\carlsim{}} class of the \texttt{pynn.carlsim} API. This technique can be used to expose other non-standard \carlsim{} methods in the \texttt{pynn.carlsim} API.

%In Section \ref{sec:results}, we show an example of simulating an application written in PyNN using the \carlsim{} simulator. 
\subsection{Using pynn-to-carlsim Interface}
\begin{figure}[h!]
	\centering
	%\vspace{-5pt}
	\centerline{\includegraphics[width=0.96\columnwidth]{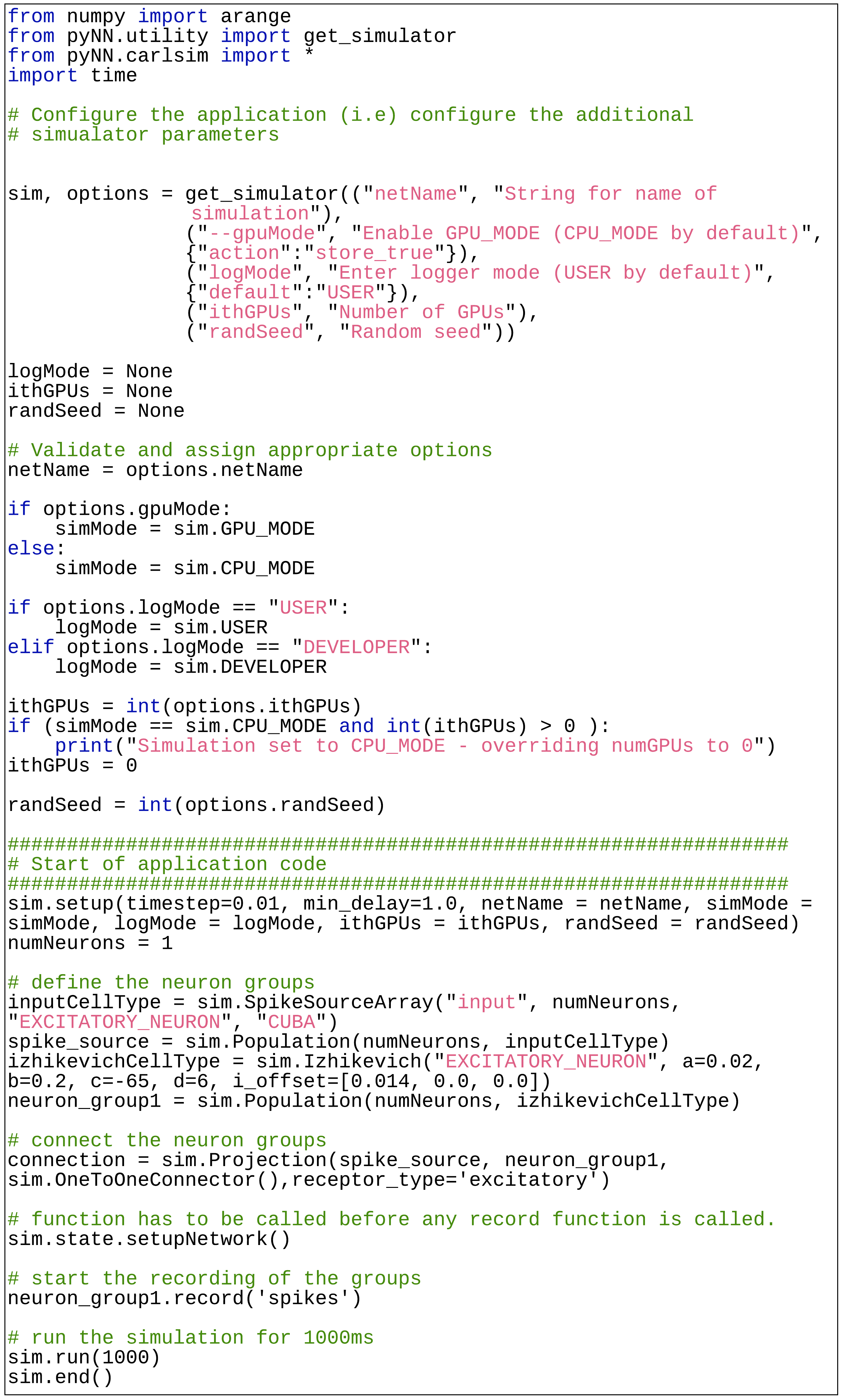}}
	\caption{An example application code written in PyNN.}
	\vspace{-10pt}
	\label{fig:pynn_carlsim_3}
\end{figure}

To verify the integrity of our implementation, Figure~\ref{fig:pynn_carlsim_3} shows the source code of a test application written in PyNN. 
The source code sets SNN parameters using PyNN. 
A
simple spike generation group with 1 neuron and a second neuron group with 3
excitatory Izhikevich neurons are created.
The user can run the code in a CPU or a GPU mode by specifying
the respective parameters in the command line.
%\hkNote{In Figure 13, the Izhikevich group has one neuron, however, you described above one-to-one connection to three neurons. What is i\_offset in the neuron group constructor, I am curious as it is a list of three floats?}

% \begin{figure}[!ht]
%      \subfloat[]{%
%       \includegraphics[width=0.99\columnwidth]{images/code_1.pdf}
%      }
%      \hfill
%      \subfloat*[]{%
%       \includegraphics[width=0.99\columnwidth]{images/code_2.pdf}
%      }
%      \caption{An example application code in PyNN.}
%      \label{fig:pynn_carlsim_1}
% \end{figure}
   
% \begin{figure}[t!]
%     \centering
%     \begin{subfigure}[t]{0.99\columnwidth}
%         \centering
%         \includegraphics[height=1.2in]{images/code_1.pdf}
%         \caption{Lorem ipsum}
%     \end{subfigure}%
%     ~ 
%     \begin{subfigure}[t]{0.99\columnwidth}
%         \centering
%         \includegraphics[height=1.2in]{images/code_2.pdf}
%         \caption{Lorem ipsum, lorem ipsum,Lorem ipsum, lorem ipsum,Lorem ipsum}
%     \end{subfigure}
%     \caption{Caption place holder}
% \end{figure}

% \captionsetup[subfigure]{labelformat=empty}
% \begin{figure}%
%     \centering
%     \subfloat[]{{\includegraphics[width=0.99\columnwidth]{images/code_1.pdf}}}%
%     \quad
%     \vspace{-30pt}
%     \subfloat[]{{\includegraphics[width=0.99\columnwidth]{images/code_2.pdf} }}%
%     \caption{2 Figures side by side}%
%     \label{fig:example}%
% \end{figure}

\vspace{-10pt}

\vspace{-10pt}

\begin{figure}[h!]
	\centering
	\vspace{-5pt}
	\centerline{\includegraphics[width=0.96\columnwidth]{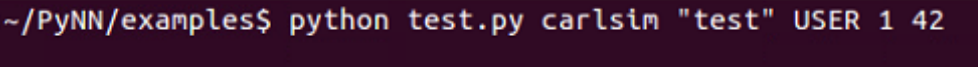}}
	\caption{Executing the test application from command line.}% using \carlsim{} as the back-end simulator.}
	\vspace{-10pt}
	\label{fig:command_line}
\end{figure}

It can be seen from the figure that command line arguments have been specified
to receive the simulator specific parameters from the user.
This command line
parameters can be replaced with custom defined methods in the \texttt{pynn.carlsim} API
of PyNN or by using configuration files in xml or json format.
The application code shows the creation of an SNN model by calling the \texttt{get\_simulator} method.
To use \carlsim{} back-end, an user must specify ``carlsim" as the simulator to
be used for this script while invoking the Python script, as can be seen in Figure \ref{fig:command_line}.
This results in the internal resolution of creating an instance of \carlsim{} internally
by PyNN. The returned object (denoted as \texttt{sim} in the figure) is then used to access all
the internal API methods offering the user control over the simulation.

% \begin{figure}[h!]
% 	\centering
% 	\vspace{-10pt}
% 	\centerline{\includegraphics[width=0.99\columnwidth]{images/code_1.pdf}}
% 	\captionsetup{labelformat=empty}
% 	%\caption{Test application code in PyNN (part 1).}
% 	\vspace{-10pt}
% 	%\label{fig:pynn_carlsim_1}
% \end{figure}

\begin{figure}[h!]
	\centering
	\vspace{-5pt}
	\centerline{\includegraphics[width=0.96\columnwidth]{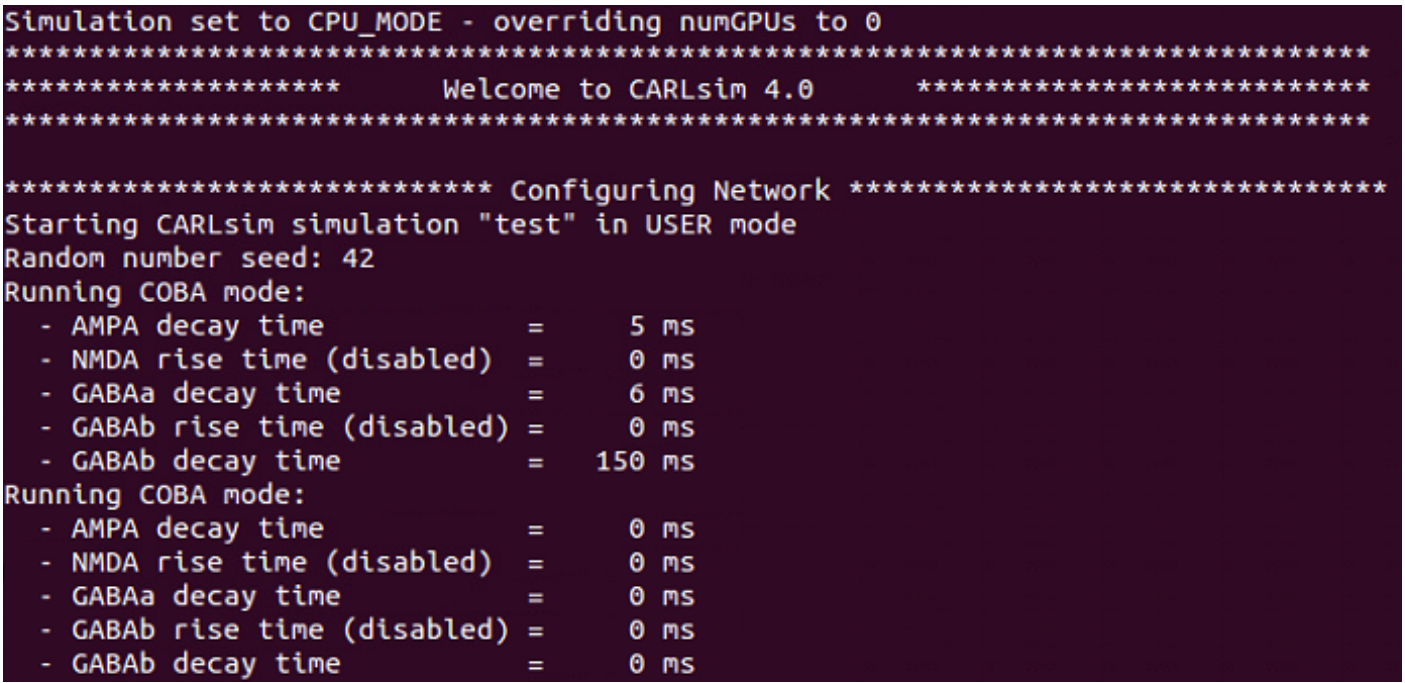}}
	\caption{Simulation snippet.}
	\vspace{-10pt}
	\label{fig:simulation_results}
\end{figure}

Figure \ref{fig:simulation_results} shows the starting of the simulation by executing the command in Figure \ref{fig:command_line}. As can be seen, the CPU\_MODE is set by overriding the GPU\_MODE as no
argument was provided in the command and the default mode being CPU\_MODE
was set. The test is being run in the logger mode USER with a random seed of 42 as specified in the command line. From Figure \ref{fig:pynn_carlsim_3}, we see that the application is set
in the current-based (CUBA) mode, which is also reported during simulation (Fig.~\ref{fig:simulation_results}). The timing parameters such as the AMPA decay time and the GABAb decay times
are set in simulation as shown in Figure \ref{fig:pynn_carlsim_3}.
%\jeffNote{The Figure 13 example is CUBA, but you describe a COBA model in the above paragraph. AMPA and GABA time constants are not set in the Izhikevich parameters. There is a separate setConductances command I believe.}

\vspace{-5pt}

\vspace{-5pt}

\vspace{-20pt}

%In Section \ref{sec:results}, we evaluate the memory and timing overhead of this proposed pynn-to-carlsim interface using both synthetic and realistic applications.

\subsection{Generating Output \textcolor{blue}{\texttt{snn.sw.out}}}
At the end of simulation, the proposed pynn-to-carlsim interface generates the following information.
\begin{itemize}
    \item \emph{\textcolor{blue}{Spike Data:}} the exact spike times of all neurons in the SNN model and stores them in a 2D spike vector. The first dimension of the vector is neuron id and the second dimension is spike times. Each element \ineq{spkVector[i]} is thus a vector of all spike times for the \ineq{i^\text{th}} neuron.
    \item \emph{\textcolor{blue}{Weight Data:}} the synaptic weights of all synapses in the SNN model and stores them in a 2D connection vector. The first dimension of the vector is the pre-synaptic neuron id and the second dimension is the post-synaptic neuron id. The element \ineq{synVector[i,j]} is the synaptic weight of the connection \ineq{(i,j)}.
\end{itemize}

The spike and weight data can be used to analyze and adjust the SNN model. They form the output \texttt{\textcolor{blue}{snn.sw.out}} of our integrated framework \tech{}.

% \begin{figure}[h!]
% 	\centering
% 	\vspace{-10pt}
% 	\centerline{\includegraphics[width=0.99\columnwidth]{images/simulation.pdf}}
% 	\caption{Simulation snippet.}
% 	%\vspace{-10pt}
% 	\label{fig:simulation_results}
% \end{figure}

%\vspace{-5pt}
\section{Hardware-Oriented Simulation in \tech{}}\label{sec:hardware_simulation}

To estimate the performance impact of executing SNNs on a neuromorphic hardware, the standard approach is to map the SNN to the hardware and measure the change in spike timings, which are then analyzed to estimate the performance deviation from software simulations. However, there are three limitations to this approach. First, neuromorphic hardware are currently in their research and development phase in a selected few research groups around the world. They are not yet commercially available to the bigger systems community. Second, neuromorphic hardware that are available for research have \textcolor{black}{limitations on the number of synapses per neuron}. For instance, DynapSE can only accommodate a maximum of 128 synapses per neuron. These hardware platforms therefore cannot be used to estimate performance impacts \textcolor{black}{on large-scale SNN models}. Third, existing hardware platforms \textcolor{black}{have limited} interconnect strategies for communicating spikes between neurons, and therefore they cannot be used to explore the design of scalable neuromorphic architectures that minimize latency, a key requirement for executing real-time machine learning applications. To address these limitations, we propose to design a cycle-accurate neuromorphic hardware simulator, which can allow the systems community to explore current and future neuromorphic hardware to simulate large SNN models and estimate the performance impact.

\vspace{-5pt}

\subsection{Designing Cycle-Accurate Hardware Simulator}
\begin{figure}[h!]
	\centering
	\vspace{-10pt}
	\centerline{\includegraphics[width=0.99\columnwidth]{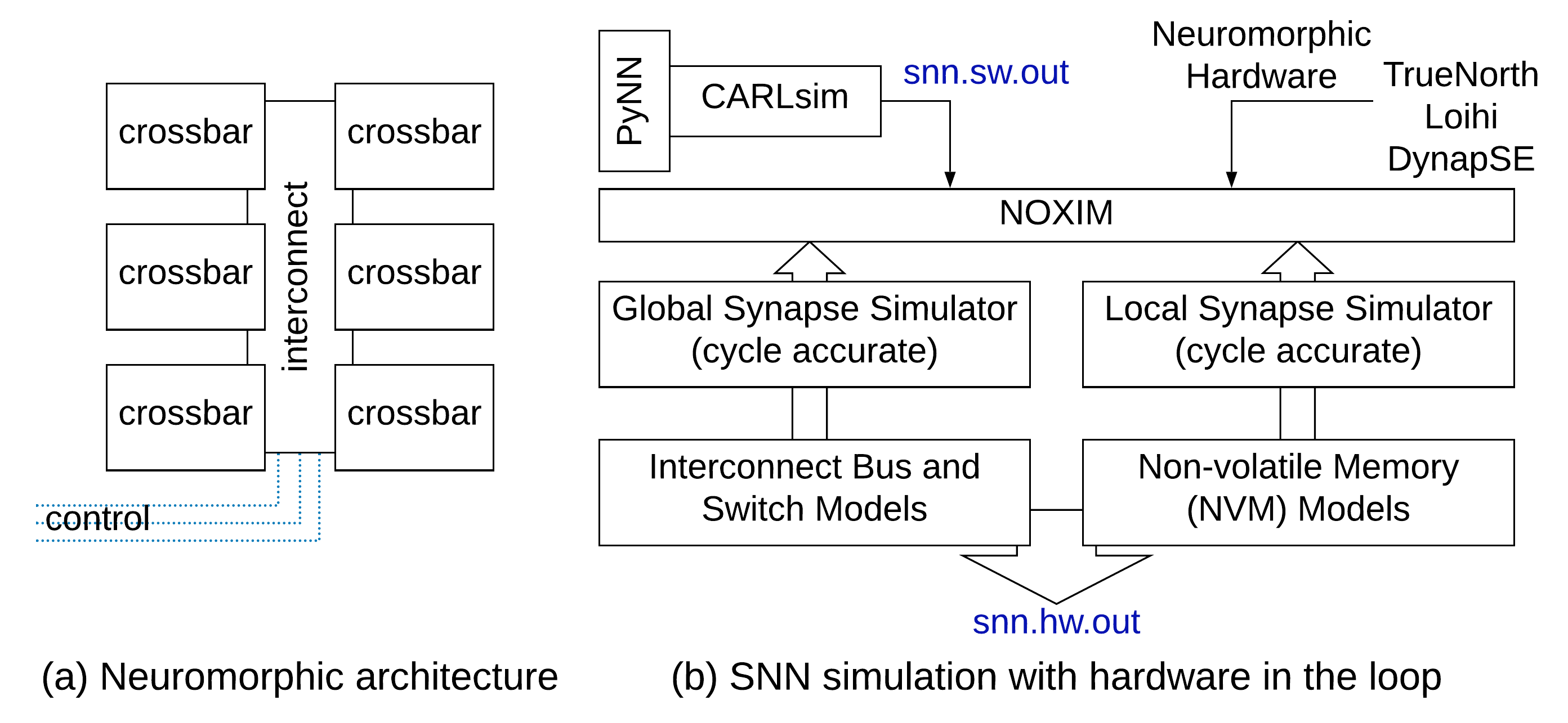}}
	\caption{(a) Neuromorphic architecture and (b) SNN Simulations with hardware in the loop.}
	\vspace{-10pt}
	\label{fig:noxim}
\end{figure}

Figure \ref{fig:noxim}(a) shows the architecture of a neuromorphic hardware with multiple crossbars and a shared interconnect.
Analogous to the mammalian brain, synapses of a SNN can be grouped into local and global synapses based on the distance information (spike) conveyed. Local synapses are short distance links, where pre- and post-synaptic neurons are located in the same vicinity. They map inside a crossbar. Global synapses are those where pre- and post-synaptic neurons are farther apart. To reduce power consumption of the neuromorphic hardware, the following strategies are adopted:
\begin{itemize}
	\item the number of point-to-point local synapses is limited to a reasonable dimension (size of a crossbar); and
	\item instead of point-to-point global synapses (which are of long distance) as found in a mammalian brain, the hardware implementation usually consists of time-multiplexed interconnect shared between global synapses.
\end{itemize}
DYNAP-SE \cite{Moradi2018ADYNAPs} for example, consists of four crossbars, each with 128 pre- and 128 post-synaptic neurons implementing a full 16K (128x128) local synapses per crossbar.

%\jeffNote{Fix (a) and (a) label in Figure.}

\vspace{-10pt}

%\jeffNote{Make the font bigger inside the boxes of fig:noxim}

Since local synapses map within the crossbar, their latency is fixed and can be estimated offline. However, the global synapses are affected by variable latency introduced due to time multiplexing of the shared interconnect at run-time.
Figure \ref{fig:noxim}(b) shows the proposed framework for SNN simulation with hardware in the loop. The \textcolor{blue}{\texttt{snn.sw.out}} generated from the pynn-to-carlsim interface is used as trace for the cycle-accurate simulator NOXIM \cite{catania2015noxim}. NOXIM allows integration of circuit-level power-performance models of non-volatile memory (NVM), e.g., phase-change memory (PCM) for the crossbars and highly configurable global synapse model based on mesh architecture. The user configurable parameters include buffer size, network size, packet size, packet injection rate, routing
algorithm, and selection strategy. 
In the power consumption simulation aspect, a user
can modify the power values in external loaded YAML file to benefit from the flexibility. For the
simulation results, NOXIM can calculate latency, throughput and power consumption automatically
based on the statistics collected during runtime.

NOXIM has been developed using a modular structure that easily allows to add new
interconnect models, which is an adoption of object-oriented programming methodology, and to experiment
with them without changing the remaining parts of the simulator code. The cycle-accurate feature is provided via the SystemC programming language. This makes NOXIM the ideal framework to represent a neuromorphic hardware. 

\subsubsection{Existing NOXIM Metrics}
As a traditional interconnect simulator, NOXIM provides performance metrics, which can be adopted to global synapse simulation directly. %This includes:
\begin{itemize}
    \item \textbf{Latency:} The difference between the sending and receiving time of spikes in number of cycles. 
    \item \textbf{Network throughput:} The number of total routed spikes divided by total simulation time in number of cycles.
    \item \textbf{Area and energy consumption:} Area consumption is calculated based on the number
of processing elements and routers; energy consumption is generated based on not only the
number, but also their activation degree depending on the traffic. The area and energy consumption are high-level estimates for a given neuromorphic hardware. We adopt such high-level approach to keep the simulation speed sufficiently low, which is required to enable
the early design space exploration.
\end{itemize}

\begin{figure*}[t!]%
	\centering
	\subfloat[][Impact of ISI distortion on accuracy. Top sub-figure shows a scenario where an output spike is generated based on the spikes received from the three input neurons. Bottom sub-figure shows a scenario where the second spike from neuron 3 is delayed. There are no output spikes  generated. \label{fig:isi_imact}]{
		\includegraphics[width=0.43\columnwidth]{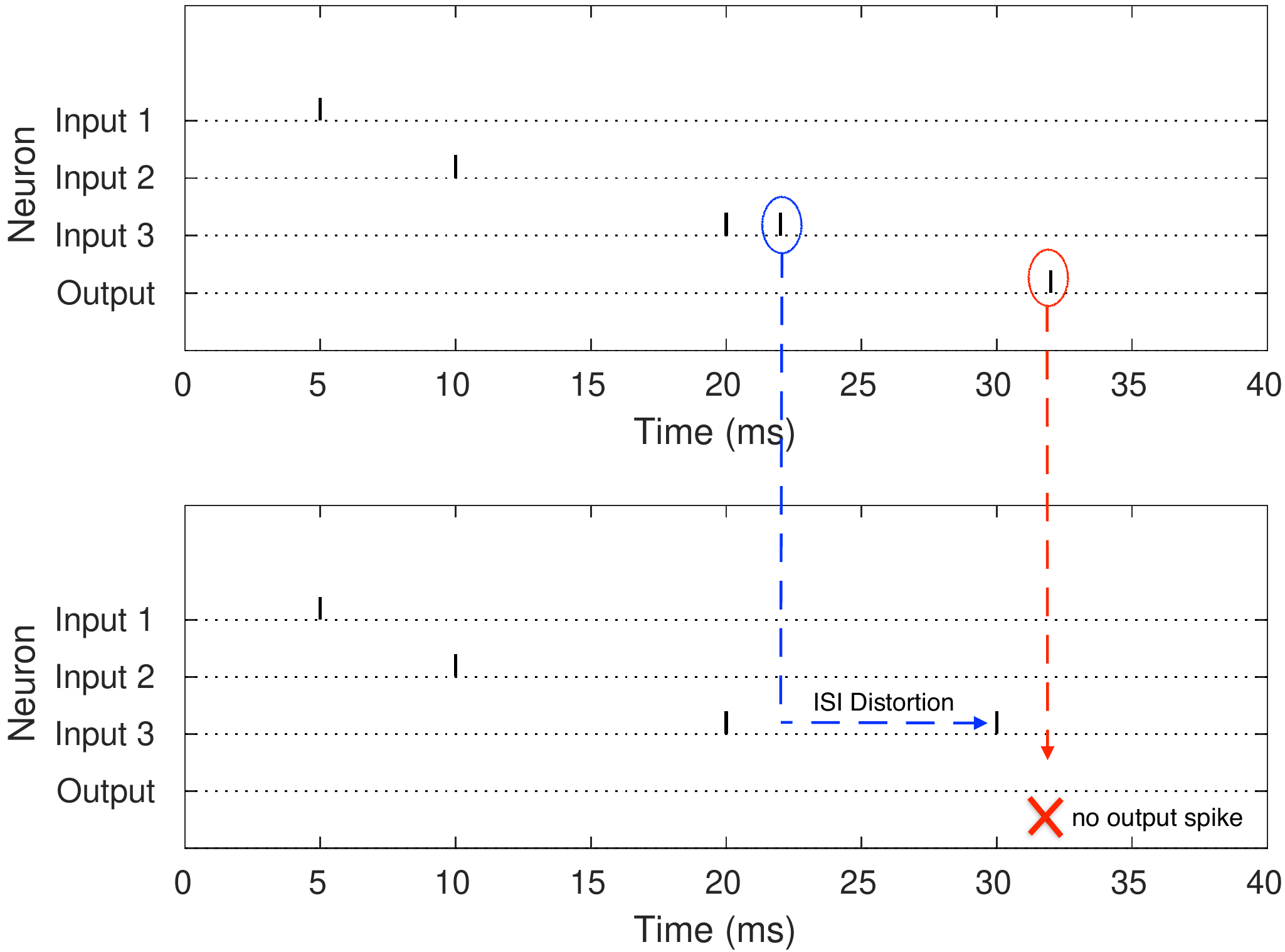}
	}
	\quad
	\subfloat[][Impact of spike disorder on accuracy. Top sub-figure shows a scenario where spike A is received at the output neuron before spike B, causing the output spike at 21ms. Bottom sub-figure shows a scenario where the spike order of A \& B is reversed. There are no output spikes  generated as a result. \label{fig:disorder_impact}]{
		\includegraphics[width=0.44\columnwidth]{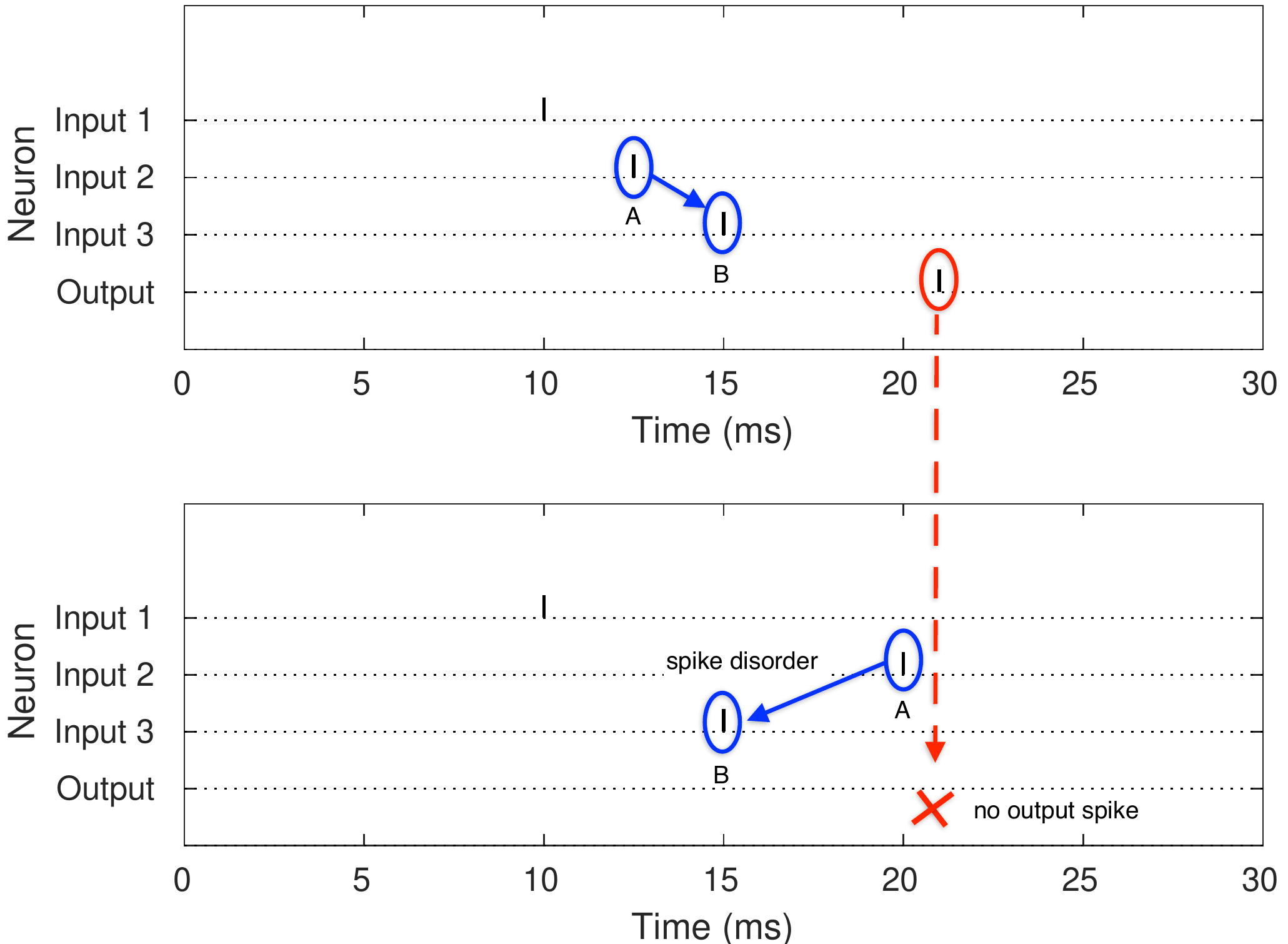}
	}
	\caption{Impact of ISI distortion (a) and spike disorder (b) on the output spike for a simple SNN with three input neurons connected to a single output neuron.}
	\label{3figs}
	\vspace{-8pt}
\end{figure*}

\subsubsection{New NOXIM Metrics}
We introduce the following two new metrics to represent the performance impact of executing an SNN on the hardware.
\begin{itemize}
    \item \textbf{Disorder spike count:} \nc{This is added for SNNs where information is encoded in terms of spike rate.
    We formulate spike disorder as follows. Let \ineq{F^i = \{F_1^i,\cdots,F_{n_i}^i\}} be the expected spike arrival rate at neuron \ineq{i}  and \ineq{\hat{F}^i = \{\hat{F}_1^i,\cdots,\hat{F}_{n_i}^i\}} be the actual spike rate considering hardware latencies. The spike disorder is computed as 
    \begin{equation}
    \label{eq:spike_disorder}
    \footnotesize \text{spike disorder} = \sum_{j=1}^{n_i} [(F_j^i - \hat{F}_j^i)^2] / n_i
    \end{equation}}
    %This is added for SNNs where information is encoded in terms of spike frequency. The dynamic spiking frequency is calculated based on the interval of two successive spikes. The difference in the spike order between a sender and a receiver measures the amount of information loss in SNNs.
    %\jeffNote{When I think of spike frequency, I think of firing rates. But, this may not be as impacted by disorder as spike timing or the order of spikes received.}
    \item \textbf{Inter-spike interval distortion:} Performance of supervised machine learning is measured in terms of \textit{accuracy}, which can be assessed from inter-spike intervals (ISIs) \cite{grun2010analysis}. To define ISI, we let \ineq{\{t_1,t_2,\cdots,t_{K}\}} be a neuron's firing times in the time interval \ineq{[0,T]}. %Neural activities in this time interval generate $K$ spikes with spike times \ineq{\{t_1,t_2,\cdots,t_{K}\}}.
The average ISI of this spike train is given by \cite{grun2010analysis}:
\begin{equation}
    \label{eq:isi}
    \vspace{-5pt}
    \footnotesize \mathcal{I} = \sum_{i=2}^K (t_i - t_{i-1})/(K-1).
\end{equation}
    % \item \textbf{Fan-out ratio:} This metric is aimed to incorporate how an SNN model is supported on a specific communication mechanism or network architecture. The fan-out ratio is defined as the total number of received spikes in all neurons divided by the total number of generated spikes in all neurons and calculated as
    % \begin{equation}
    %     \label{eq:fo}
    %     \footnotesize \text{fan-out} = \frac{N_\text{total received spikes}}{N_\text{total generated spikes}}
    % \end{equation}
\end{itemize}

%\vspace{-10pt}

To illustrate how ISI distortion and spike disorder impact accuracy, we consider a small SNN example where three input neurons are connected to an output neuron. In Figure \ref{fig:isi_imact}, we illustrate the impact of ISI distortion on the output spike. In the top sub-figure, we observe that a spike is generated at the output neuron at 22ms due to spikes from the input neurons. In the bottom sub-figure, we observe that the second spike from input 3 is delayed, i.e., has ISI distortion. As a result of this distortion, there is no output spike. Missing spikes can impact application accuracy, as spikes encode information in SNNs. In Figure \ref{fig:disorder_impact}, we illustrate the impact of spike disorder on the output spike. In the top sub-figure, we observe that the spike A from input 2 is generated before the spike B from input 3, causing an output spike to be generated at 21ms. 
In the bottom sub-figure, we observe that the spike order of inputs 2 and 3 is reversed, i.e., the spike B is generated before the spike A. This spike disorder results in no spike being generated at the output neuron, which can also lead to a drop in accuracy.

\begin{figure}[h!]
	\centering
	\vspace{-5pt}
	\centerline{\includegraphics[width=0.7\columnwidth]{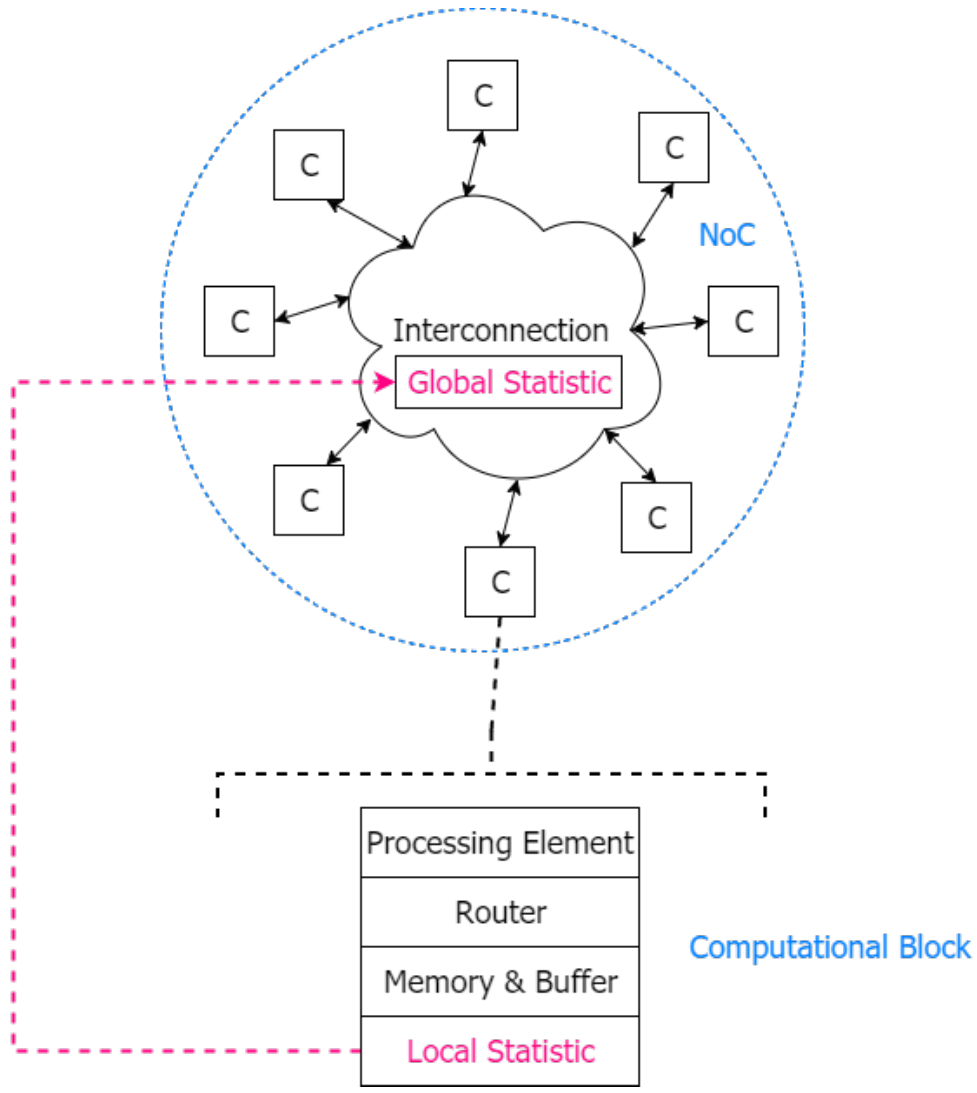}}
	\caption{Statistics collection architecture in \tech{}.}
	\vspace{-10pt}
	\label{fig:stat}
\end{figure}

\subsection{Generating Output \textcolor{blue}{\texttt{snn.hw.out}}}
Figure \ref{fig:stat} shows the statistics collection architecture in \tech{}. 
Overall, the output \textcolor{blue}{\texttt{snn.hw.out}} consists of two performance components as highlighted in Table \ref{tab:snn_hw_out}.

\begin{table}[h!]
\setlength{\tabcolsep}{10pt}
\renewcommand{\arraystretch}{1.0}
\centering
{\fontsize{8}{10}\selectfont
\begin{tabu}{c p{5cm}}
    \tabucline[2pt]{-}
    \multicolumn{2}{c}{\textcolor{blue}{\texttt{snn.hw.out}}}\\
    \hline
    \multirow{2}{*}{hardware performance} & specific to neuromorphic hardware\\\cline{2-2}
    & latency, throughput, and energy\\
    \hline
    \multirow{2}{*}{model performance} & specific to SNN model\\\cline{2-2}
    & disorder, inter-spike interval, and fanout\\
    % \multirow{3}{*}{\textbf{Operating mode}} & \multicolumn{3}{c}{\textbf{Operating voltage}} \\ \cline{2-4}
    % & \textbf{pulse shaper} & \textbf{verify} & \textbf{sense amplifier} \\
    % & \textbf{(PS)} & \textbf{(VF)} & \textbf{(SA)}\\
    % \hline
    % Read & 1.2V & 1.2V & 2.85V\\
    % Write (program) & 3.7V & 2.85V & 1.2V\\
    % Idle & 1.2V & 1.2V & 1.2V\\
    % %\hline
    % De-stress & $<V_\text{th}$ & $<V_\text{th}$ & $<V_\text{th}$\\
    \tabucline[2pt]{-}
\end{tabu}
}
%\vspace{-10pt}
\caption{Performance metrics obtained in executing an SNN model on the neuromorphic hardware.}
\label{tab:snn_hw_out}
\end{table}

% \section{Hardware Mapping using \tech{}}\label{sec:hardware_mapping}
% \input{ijcnn2020/sections/spinemap.tex}

\section{Evaluation Methodology}\label{sec:evaluation_methodology}
\subsection{Simulation Environment}
We conduct all experiments on a system with 8 CPUs, 32GB RAM, and NVIDIA Tesla GPU, running Ubuntu 16.04. 

\subsection{Evaluated Applications}
Table \ref{tab:apps} reports the applications that we used to evaluate \tech{}. The application set consists of 7 functionality tests from the \carlsim{} and PyNN repositories. The \carlsim{} functionality tests are testKernel\{1,2,3\}. The PyNN functionality tests are Izhikevich, Connections, SmallNetwork, and Varying\_Poisson. These functionalities verify the biological properties on neurons and synapses. Columns 2, 3 and 4 in the table reports the number of synapses, the SNN topology and the number of spikes simulated by these functionality tests.

Apart from the functionality tests, we evaluate \tech{} using large SNNs for 4 synthetic and 4 realistic applications.
The synthetic applications are indicated with the letter `S' followed by a number (e.g., S\_1000), where the number represents the total number of neurons in the application. %Column 3 reports the number of synapses in these applications.  Column 4 reports the SNN topology. 
The 4 realistic applications are \textit{image smoothing} (ImgSmooth) \cite{chou2018carlsim} on 64x64 images, \textit{edge detection} (EdgeDet) \cite{chou2018carlsim} on 64x64 images using difference-of-Gaussian, \textit{multi-layer perceptron (MLP)-based handwritten digit recognition} (MLP-MNIST) \cite{diehl2015unsupervised} on 28x28 images of handwritten digits and \textit{CNN-based heart-beat classification} (HeartClass) using ECG signals \cite{balaji2018power,das2018unsupervised,das2018heartbeat}. 
% , \textit{CNN-based digit classification} (CNN-MNIST) \cite{springenberg2014striving,mlperf}, \textit{CNN-based digit classification with LeNet} (LeNet-MNIST) \cite{mlperf}, and \textit{CNN-based CIFAR image classification with LeNet} (LeNet-CIFAR) \cite{mlperf}.
% The last three applications are part of the MLPerf benchmark suite \cite{mlperf} and developed for analog computation model. We converted these applications into spike-based model using the CNN-to-SNN conversion tool N2D2 \cite{n2d2,diehl2016conversion}.

\begin{table}[h!]
	\renewcommand{\arraystretch}{0.8}
	\setlength{\tabcolsep}{2pt}
	\centering
	\begin{threeparttable}
	{\fontsize{6}{10}\selectfont
		\begin{tabular}{cc|cl|c}
			\hline
			\textbf{Category} & \textbf{Applications} & \textbf{Synapses} & \textbf{Topology} & \textbf{Spikes}\\
			\hline
			\multirow{7}{*}{functionality tests} & testKernel1 & 1 & FeedForward (1, 1) & 6\\
			& testKernel2 & 101,135 & Recurrent (Random) & 96,885\\
			& testKernel3 & 100,335 & FeedForward (800, 200) & 63,035\\
			& Izhikevich & 4 & FeedForward (3, 1) & 3\\
			& Connections & 7,200 & Recurrent (Random) & 1,439\\
			& SmallNetwork & 200 & FeedForward (20, 20) & 47\\
			& Varying\_Poisson & 50 & FeedForward (1, 50) & 700\\
			
			\hline
			\multirow{4}{*}{synthetic} & S\_1000 & 240,000 & FeedForward (400, 400, 100) & 5,948,200\\
			& S\_1500 & 300,000 & FeedForward (500, 500, 500) & 7,208,000\\
			& S\_2000 & 640,000 & FeedForward (800, 400, 800) & 45,807,200\\
			& S\_2500 & 1,440,000 & FeedForward (900, 900, 700) & 66,972,600\\
			\hline
			\multirow{4}{*}{realistic} & ImgSmooth \cite{chou2018carlsim} & 136,314 & FeedForward (4096, 1024) & 17,600\\
			& EdgeDet \cite{chou2018carlsim} & 272,628 &  FeedForward (4096, 1024, 1024, 1024) & 22,780\\
			& MLP-MNIST \cite{diehl2015unsupervised} & 79,400 & FeedForward (784, 100, 10) & 2,395,300\\
			& HeartClass \cite{balaji2018power} & 2,396,521 & CNN\tnote{1} & 1,036,485\\
			%Input(32x32x3) - [Conv, Pool]*6 - [Conv, Pool]*6 - FC*84 - FC*10
			\hline
	\end{tabular}}
	%}
	\begin{tablenotes}\scriptsize
        \item[1.] Input(82x82) - [Conv, Pool]*16 - [Conv, Pool]*16 - FC*256 - FC*6
        %\item[2.] Input(24x24) - [Conv, Pool]*16 - FC*150 - FC*10
        %\item[3.] Input(32x32) - [Conv, Pool]*6 - [Conv, Pool]*16 - Conv*120 - FC*84 - FC*10
        %\item[4.] Input(32x32x3) - [Conv, Pool]*6 - [Conv, Pool]*6 - FC*84 - FC*10
    \end{tablenotes}
	\end{threeparttable}
	\caption{Applications used for evaluating \tech{}.}
	\label{tab:apps}
\end{table}

\vspace{-10pt}

\section{Results and Discussion}\label{sec:results}
\subsection{Evaluating pynn-carlsim Interface of \tech{}}
We evaluate the proposed pynn-to-carlsim interface in \tech{} using the following two performance metrics.
\begin{itemize}
    \item \textbf{Memory usage:} This is the amount of main memory (DDDx) occupied by each application when simulated using \tech{} (Python). Main memory usage is reported in terms of the resident set size (in kB). Results are normalized to the native \carlsim{} simulation (in C++).
    \item \textbf{Simulation time:} This is the time consumed to simulate each application using \tech{}. Execution time is measured as CPU time (in ms). Results are normalized to the native \carlsim{} simulation.
\end{itemize}

\subsubsection{\underline{Memory Usage}}
\begin{figure}[h!]
	\centering
	\vspace{-6pt}
	\centerline{\includegraphics[width=0.99\columnwidth]{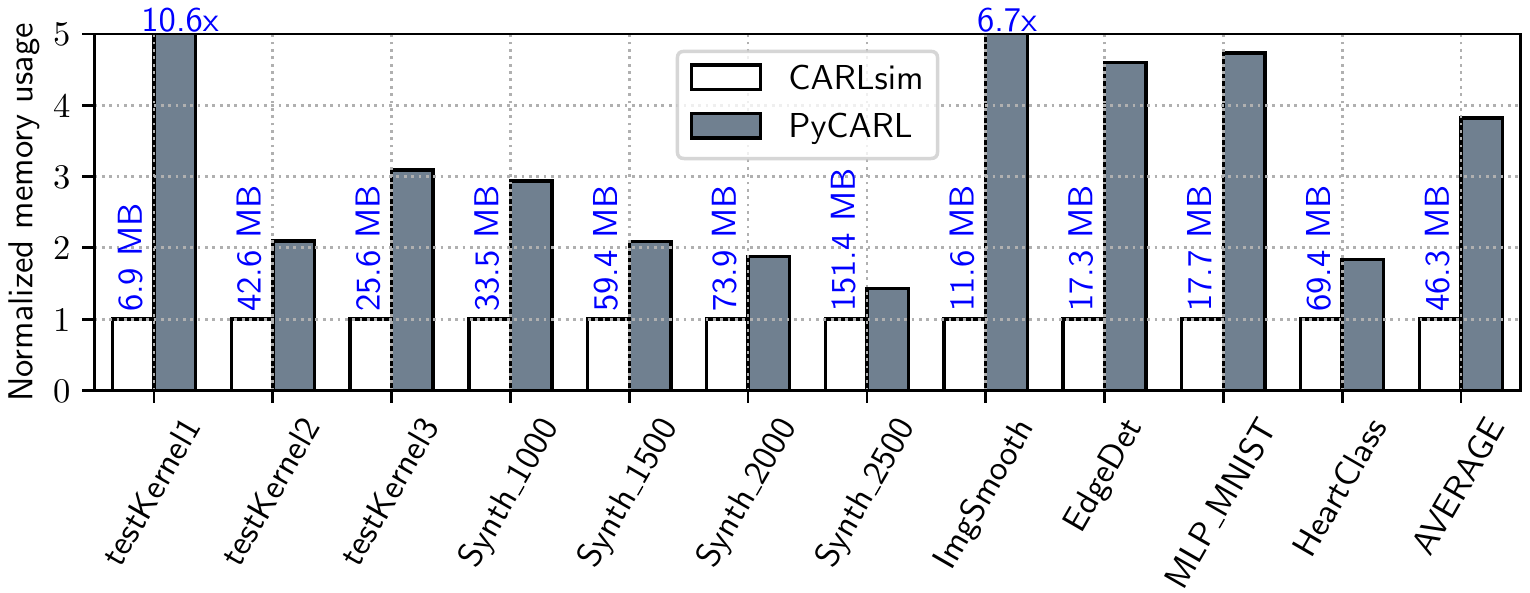}}
	\caption{Memory usage of \tech{} normalized to \carlsim{} (out of scale results are reported on the bar).}
	\vspace{-10pt}
	\label{fig:rss}
\end{figure}

Figure~\ref{fig:rss} plots the memory usage of each application using \tech{} normalized to \carlsim{}. \nc{For easy reference, the absolute memory usage of \carlsim{} is also reported on the bar for each application.} We make the following three main observations.
\nc{
First, the memory usage is application-dependent. The memory usage of testKernel1 with a single synapse is 6.9MB, compared to the memory usage of 151.4 MB for Synth\_2500 with 1,440,000 synapses.}
Second, the memory usage of \tech{} is on average 3.8x higher than \carlsim{}. This is because 1) the pynn-carlsim interface loads all shared \carlsim{} libraries in the main memory during initialization, irrespective of whether or not they are utilized during simulation and 2) some of \carlsim{}'s dynamic data structures are re-created during SWIG compilation as SWIG cannot access these native data structures in the main memory. Our future work involves solving both these limitation to reduce the memory footprint of \tech{}. Third, smaller SNNs result in higher memory overhead. This is because for smaller SNNs, the memory allocation for \carlsim{} libraries becomes the primary contributor of the memory overhead in \tech{}. \carlsim{}, on the other hand, loads only the libraries that are needed for the SNN simulation. 

%First, execution time of \static{} is lower than Baseline by an average of 5\% due to its higher de-stress interval (\ineq{tDSI}), which improves performance by reducing the de-stress overhead (Equation \ref{eq:destress_overhead}). 
% First, execution time of \tech{} is lower than Baseline by an average of 12\%. This improvement is because \tech{}, which  
% opportunistically de-stresses a peripheral circuit only when its aging exceeds the aging threshold, has a lower de-stress overhead than Baseline, which uses a fixed de-stress interval of 1000 cycles irrespective of the aging.
% %tracks the CMOS transistor aging in the peripheral circuit of each memory bank, de-stressing it only when it is absolutely necessary to improve lifetime reliability. In contrast, Baseline uses a fixed de-stress interval of 1000 cycles and therefore initiates de-stress operations even when the CMOS aging is below the aging threshold, incurring significant de-stress overhead. 
% Second, execution time of Decoupled-\tech{} is lower than \tech{} by an average of 6\%. This improvement is because Decoupled-\tech{} de-stresses a peripheral circuit off the critical path of execution and in parallel to regular read or write requests from the bank, reducing bank occupancy and improving performance.

\vspace{-10pt}

\subsubsection{\underline{Simulation Time}}
\begin{figure}[h!]
	\centering
	\vspace{-6pt}
	\centerline{\includegraphics[width=0.99\columnwidth]{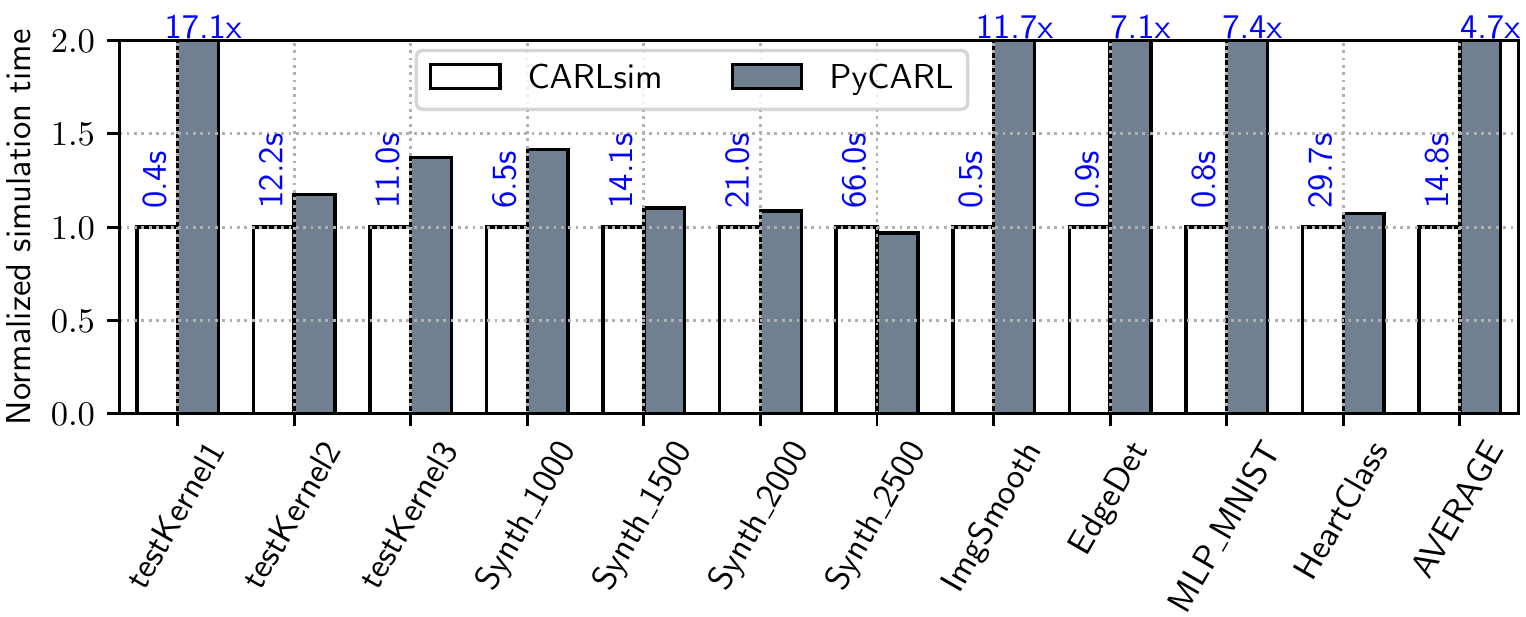}}
	\caption{Simulation time of \tech{} normalized to \carlsim{} (out of scale results are reported on the bar).}
	\vspace{-10pt}
	\label{fig:perf}
\end{figure}

%\hkNote{I am not sure if this should be called Simulation time. If we are simulating 20 timesteps, then simulation time in PyCARL and CARLsim are the same. The wall clock time maybe called execution time/runtime.
%Also, what do you mean by simpler SNN?}
Figure \ref{fig:perf} plots the simulation time of each our applications using \tech{}, normalized to \carlsim{}. \nc{For easy reference, the absolute simulation time of \carlsim{} is also reported on the bar for each application.} We make the following two main observations.
First, 
%We observe that 
the simulation time using \tech{} is on average 4.7x higher than \carlsim{}. The high simulation time of \tech{} is contributed by two components -- 1) initialization time, which includes the time to load all shared libraries and 2) the time for simulating the SNN. We observe that the simulation time of the SNN is comparable between \tech{} and \carlsim{}. The difference is in the initialization time of \tech{}, which is higher than \carlsim{}. 
Second,
%We further observe that
the overhead for smaller SNNs (i.e., ones with less number of spikes) are much higher because the initialization time dominates the overall simulation time for these SNNs, Therefore, \tech{}, which has higher initialization time, has higher simulation time than \carlsim{}.

\begin{figure}[h!]
	\centering
	%\vspace{-6pt}
	\centerline{\includegraphics[width=0.99\columnwidth]{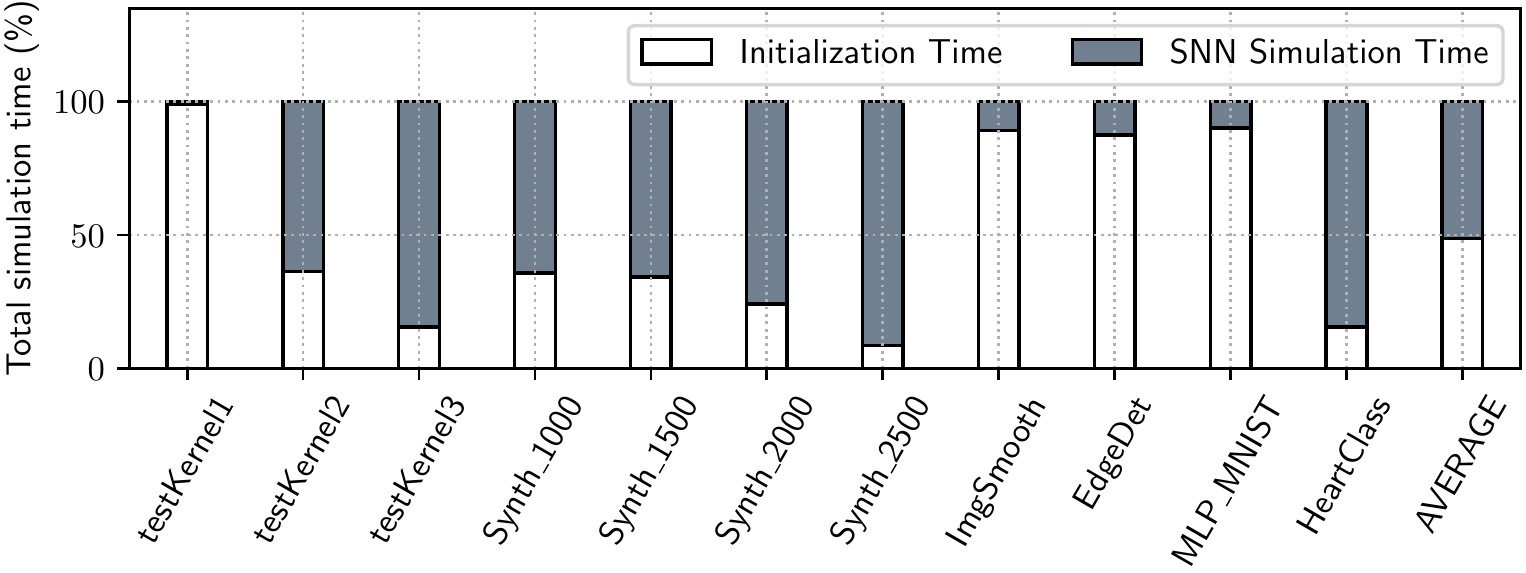}}
	\caption{Total Simulation time distributed into initialization time and SNN simulation time.}
	%\vspace{-10pt}
	\label{fig:dist}
\end{figure}

To analyze the simulation time, Figure \ref{fig:dist} plots the distribution of total simulation time into initialization time and the SNN simulation time. For testKernel1 with only 6 spikes (see Table~\ref{tab:apps}), the initialization time is over 99\% of the total simulation time. Since the initialization time is considerably higher in \tech{}, the overall simulation time is 17.1x than \carlsim{} (see Figure \ref{fig:perf}). On the other hand, for a large SNN like Synth\_2500, the initialization time is only 8\% of the total simulation time. For this application \tech{}'s simulation time is only 4\% higher than \carlsim{}.

%\vspace{-10pt}

\textit{We conclude that the total simulation time using \tech{} is only marginally higher than native \carlsim{} for larger SNNs, which are typical in most machine learning models.}
This is an important requirement to enable fast design space exploration early in the model development stage. Hardware-aware circuit-level simulators are much slower and have large memory footprint. Finally, other PyNN-based SNN simulators don't have the hardware information so
they can only provide functional checking of machine learning models.

%\jeffNote{What do the numbers above some of these bars mean? Are you cutting off the bar?  I wouldn't say just looking at the figures that it is only "marginally" higher.}

\subsection{Evaluating carlsim-hardware Interface of \tech{}}
\subsubsection{\underline{Hardware Configurations Supported in \tech{}}}
Table \ref{tab:global_networks} reports the supported spike routing algorithms 
%on global synapses supported 
in \tech{}.

\begin{table}[h!]
	\renewcommand{\arraystretch}{0.8}
	\setlength{\tabcolsep}{2pt}
	\centering
	\begin{threeparttable}
	{\fontsize{6}{10}\selectfont
		\begin{tabular}{c|p{7cm}}
			\hline
			\textbf{Algorithms} & \textbf{Description}\\
			\hline
			XY & Packets first go horizontally and then vertically to reach destinations.\\
			West First & West direction should be taken first if needed in the proposed route to destination.\\
			North Last & North direction should be taken last if needed in the proposed route to destination.\\
			Odd Even & Turning from the east at tiles located in even columns and turning to the west at tiles in odd column are prohibited.\\
			DyAD & XY routing when there is no congestion, and Odd Even routing when there is congestion.\\
			\hline
	\end{tabular}}
	%}
	\end{threeparttable}
	\caption{Routing algorithms supported in \tech{}.}
	\label{tab:global_networks}
\end{table}

\begin{figure}[h!]%
	\centering
	\subfloat[][Latency distribution. \label{fig:dyad_latency}]{
		\includegraphics[width=0.95\columnwidth]{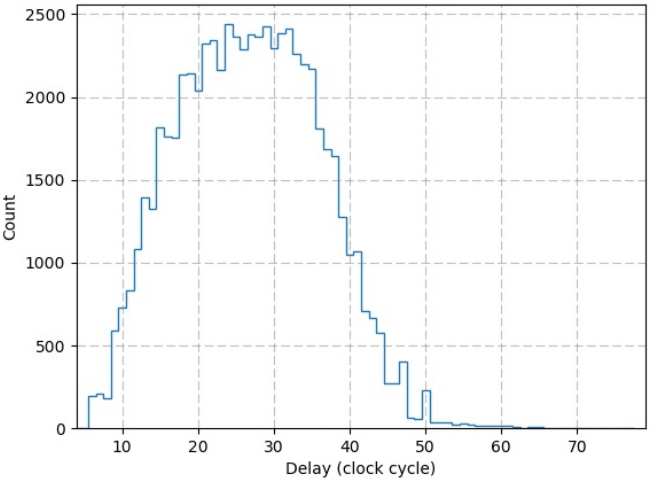}
	}
	\quad
	\subfloat[][ISI distribution. \label{fig:dyad_isi}]{
		\includegraphics[width=0.95\columnwidth]{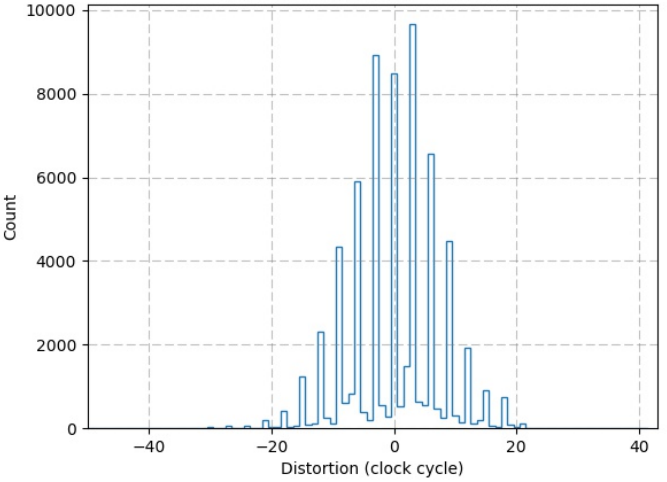}
	}
	\caption{(a) Latency and (b) ISI distortion for XY routing.}
	\label{fig:routing_algorithms}
\end{figure}

%\subsubsection{Hardware statistics collection in \tech{}} 
To illustrate the statistics collection, we use a fully connected synthetic SNN with two feedforward layers of 18 neurons each. The SNN is mapped to a hardware with 36 crossbars arranged in a 6x6 mesh topology. Figure \ref{fig:routing_algorithms} shows a typical distribution of spike latency and ISI distortion (in clock cycles) collected when configuring the global synapse network with XY routing.

Table \ref{tab:routing_algorithms} reports the statistics collected for different routing algorithms for the global synapse network of the neuromorphic hardware. \tech{} facilitates system design exploration in the following two ways. First, system designers can 
explore these statistics and set a network configuration to achieve the desired optimization objective. In our prior work \cite{balaji2019exploration}, we have developed segmented bus interconnect for neuromorphic hardware using \tech{}. Second, system designers can analyze these statistics for a given hardware to estimate performance of SNNs on hardware. In our prior work \cite{das2018mapping,balaji2019mapping,song2020compiling,balaji2019framework}, we have analyzed such performance deviation using \tech{}.

\begin{table}[h!]
	\renewcommand{\arraystretch}{0.8}
	\setlength{\tabcolsep}{2pt}
	\centering
	{\fontsize{6}{10}\selectfont
		\begin{tabular}{c|c c c c}
			\hline
			\multirow{2}{*}{\textbf{Algorithms}} & \textbf{Avg. ISI} & \textbf{Disorder Count} & \textbf{Avg. Latency} &\textbf{Avg. Throughput}\\
			& \textbf{(cycles)} & \textbf{(cycles)} & \textbf{(cycles)} & \textbf{(spikes/cycle)} \\
			\hline
			XY & 48 & 203 & 26.75 & 0.191\\
			West First & 44 & 198 & 26.76 & 0.191\\ 
			North Last & 43 & 185 & 26.77 & 0.191\\
			Odd Even & 44 & 176 & 26.77 & 0.191\\
			DyAD & 44 & 186 & 26.78 & 0.191\\
			\hline
	\end{tabular}}
	%}
	\caption{Evaluating routing algorithms in \tech{}.}
	\label{tab:routing_algorithms}
\end{table}

%\subsubsection{Software-only simulation vs. hardware-software co-simulation}
\subsubsection{\underline{Performance Impact on Hardware}}
\begin{figure}[h!]
	\centering
	%\vspace{-10pt}
	\centerline{\includegraphics[width=0.99\columnwidth]{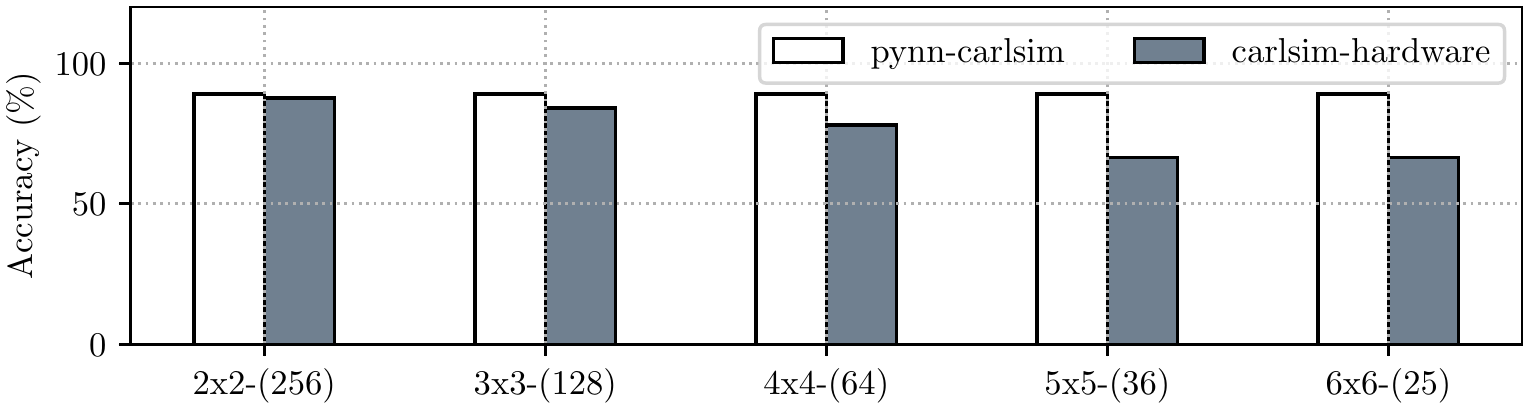}}
	\caption{Accuracy of MLP\_MNIST on five neuromorphic hardware configurations compared to the accuracy obtained via software-only simulation using \texttt{pynn-carlsim} interface.}
	\vspace{-10pt}
	\label{fig:accuracy_impact}
\end{figure}

%\subsubsection{Analyzing software-only simulation performance vs. hardware-software co-simulation performance} 
\nc{To illustrate how the performance of a machine learning application changes on hardware, 
Figure \ref{fig:accuracy_impact} shows the accuracy of MLP\_MNIST obtained on five hardware configurations programmed in \tech{}. We also report the accuracy of MLP\_MNIST obtained using software-only simulation with the proposed \texttt{pynn-carlsim} interface. The hardware configuration \ineq{n\times n~(m)} is for a neuromorphic hardware with \ineq{n^2} crossbars, arranged using a \ineq{n\times n} mesh network. Each crossbar can accommodate \ineq{m} input and \ineq{m} output neurons, with a maximum of \ineq{m} pre-synaptic connections per output neuron.} We observe that compared to an accuracy of 89\% obtained using the \texttt{pynn-carlsim} interface, the best case accuracy on a \ineq{6 \times 6} hardware (\ineq{36} crossbars with \ineq{25} input and \ineq{25} output neurons per crossbar) is only 66.6\% -- a loss of \ineq{22.4\%}. This loss is due to hardware latencies, which delay some spikes more than others, and are not accounted when performing accuracy estimation through software-only simulations. \nc{The proposed \texttt{carlsim-hardware} interface in \tech{} facilitates estimating accuracy (performance in general) impact of machine learning applications on neuromorphic hardware.}%, early in the development stage.

\subsubsection{\underline{SNN Performance on DYNAP-SE}}
\begin{figure}[h!]
	\centering
	\vspace{-6pt}
	\centerline{\includegraphics[width=0.99\columnwidth]{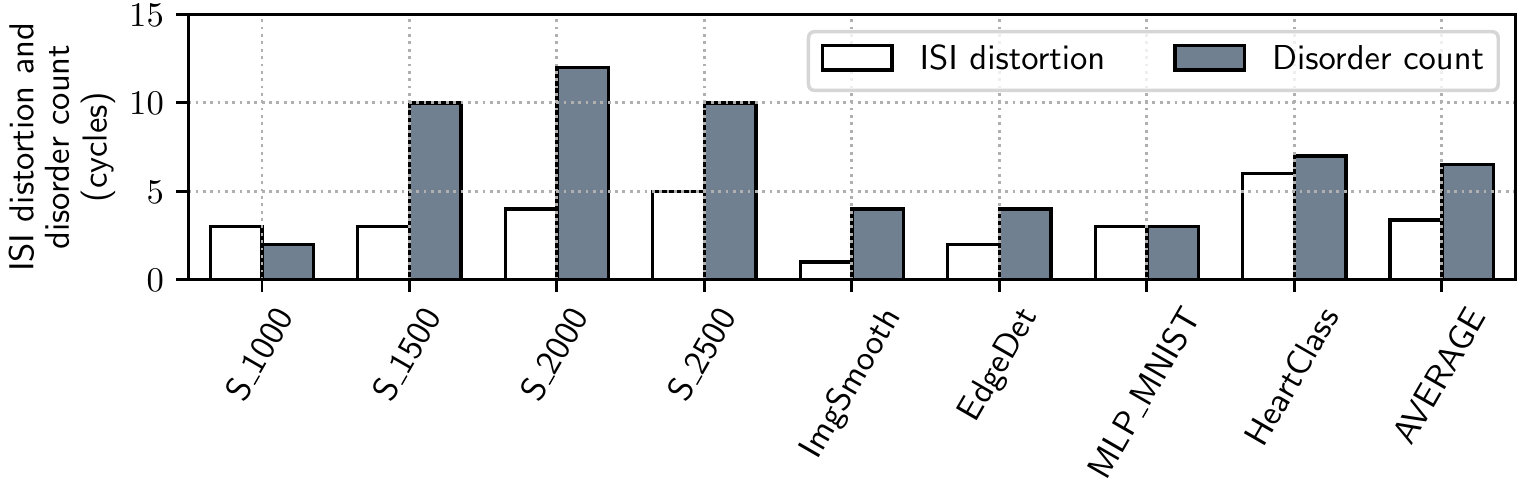}}
	\caption{ISI distortion and disorder of hardware-oriented simulation, normalized to software-only simulation.}
	\vspace{-10pt}
	\label{fig:isi_dis}
\end{figure}

%\subsubsection{Evaluating \tech{} on DynapSE} 
\nc{Figure \ref{fig:isi_dis} evaluates the statistics collection feature of \tech{} on DYNAP-SE~\cite{Moradi2018ADYNAPs}, a state-of-the-art neuromorphic hardware to estimate performance impact between software-only simulation (using \texttt{pynn-carlsim} interface) and hardware-oriented simulation (using \texttt{carlsim-hardware} interface) for each application.}

We observe that these applications have average ISI distortion of 3.375 cycles and disorder of 6.5 cycles when executed on the specific neuromorphic hardware. \nc{In the software-only simulation (using the \texttt{pynn-carlsim} interface), ISI distortion and disorder count are both zero.} These design metrics directly influence performance, as illustrated in Figure \ref{3figs}.

\subsubsection{\underline{Design Space Exploration using \tech{}}} 
%Recently, researchers have demonstrated using PyNN to map machine learning applications to neuromorphic hardware such as SpiNNaker, 
% %and 
%BrainScaleS, and 
% %more recently to Intel 
%Loihi by balancing the number of synapses on different neurosynaptic cores of the hardware. 
We now demonstrate how the statistics collection feature of \tech{} can be used to perform design space explorations 
%on DynapSE \cite{Moradi2018ADYNAPs}, a state-of-the-art neuromorphic hardware, 
optimizing 
%other 
hardware metrics such as latency and energy.
%1) hardware performance measures such energy, latency, and throughput, and 2) application performance metrics such as ISI distortion and disorder count.
We demonstrate \tech{} for DYNAP-SE 
%\cite{Moradi2018ADYNAPs}, a state-of-the-art neuromorphic hardware 
using an instance of particle swarm optimization (PSO) \cite{kennedy1995particle} to distribute the synapses in order to minimize latency and energy. The mapping technique is adapted from our earlier published work \cite{balaji2019mapping}. Although optimizing SNN mapping to the hardware is not the main focus of this
paper, the following results only illustrate the capability of
\tech{} to perform such optimization.
%Our framework can be modified with minimal efforts to model other neuromorphic hardware.

\begin{figure}[h!]
	\centering
	\vspace{-6pt}
	\centerline{\includegraphics[width=0.99\columnwidth]{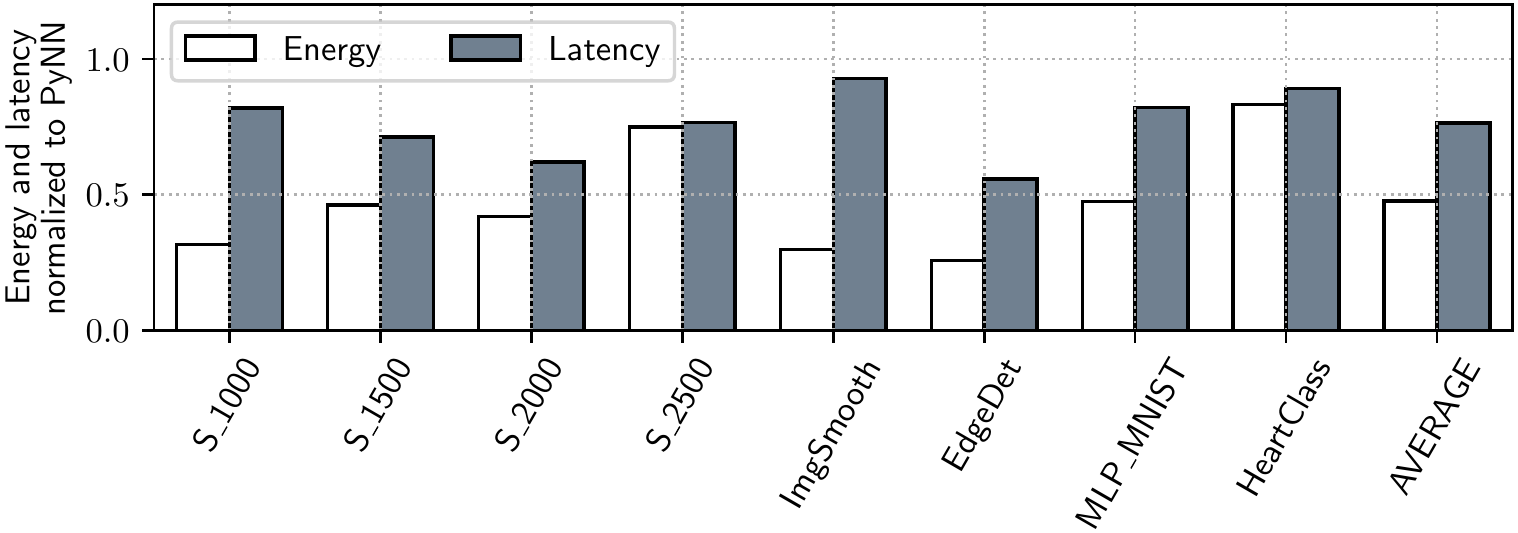}}
	\caption{Energy and latency of \tech{} normalized to PyNN.}
	\vspace{-10pt}
	\label{fig:en_lat}
\end{figure}

Figure \ref{fig:en_lat} plots the energy and latency of each application obtained using \tech{}, normalized to PyNN, which balances the synapses on different crossbars of the hardware. We observe that \tech{} achieves an average 50\% lower energy and 24\% lower latency than PyNN's native load balancing strategy. These improvements clearly motivate the significance of \tech{} in advancing neuromorphic computing.
%DynapSE consists of four neurosynaptic cores (i.e., crossbars) integrated using a two-stage networks-on-chip (NoC). We illustrate 

\vspace{-10pt}

%\section{Related Works}\label{sec:related_works}

\section{Conclusions}
\label{sec:conclusion}
We present \tech{}, a 
Python programming interface that allows \carlsim{}-based spiking neural network simulations with neurobiological details at the neuron and synapse levels, hardware-oriented simulations, and design-space explorations for neuromorphic computing, all from a common PyNN frontend. 
\tech{} allows extensive portability across different research institutes. We evaluate \tech{} using functionality tests as well as synthetic and realistic SNN applications on a state-of-the-art neuromorphic hardware.
By using cycle-accurate models of neuromorphic hardware, \tech{} allows users to perform neuromorphic hardware and machine learning model explorations and performance estimation
%, without the necessity to procure these hardware platforms 
early during application development, accelerating the neuromorphic product development cycle.
We \textbf{conclude} that \tech{} is a comprehensive framework that has significant potential to advance the field of neuromorphic computing.

\nc{\tech{} is available for download at \cite{pycarl}.} %framework that has significant potential to advance the field of neuromorphic computing.

\section*{Acknowledgment}
This work is supported by 1) the National Science Foundation Award CCF-1937419 (RTML: Small: Design of System Software to Facilitate Real-Time Neuromorphic Computing) and 2) the National Science Foundation Faculty Early Career Development Award CCF-1942697 (CAREER: Facilitating Dependable Neuromorphic Computing: Vision, Architecture, and Impact on Programmability).

%\section*{References}
\bibliographystyle{IEEEtran}
\bibliography{snn_tool}

\end{document}